\documentclass[journal]{IEEEtran}
\usepackage{amsmath,amsfonts}
\usepackage{algorithmic}
\usepackage{algorithm}
\usepackage{array}
\usepackage[caption=false,font=normalsize,labelfont=sf,textfont=sf]{subfig}
\usepackage{textcomp}
\usepackage{stfloats}
\usepackage{url}
\usepackage{verbatim}
\usepackage{graphicx}
\usepackage{cite}
\usepackage{color}
\usepackage{hyperref}
\hypersetup{
    colorlinks=true,
    linkcolor=blue,
    filecolor=blue,      
    urlcolor=blue,
    citecolor=blue,
}
\begin{document}
\title{Decoding Modular Reconfigurable Robots: \\A Survey on Mechanisms and Design
}

\author{Guanqi~Liang,~\IEEEmembership{Student Member,~IEEE},
        Di~Wu,~\IEEEmembership{Student Member,~IEEE},
        
        Yuxiao~Tu,~\IEEEmembership{Student Member,~IEEE},
        and~Tin Lun~Lam,~\IEEEmembership{Senior Member,~IEEE}

\thanks{This work was supported in part by the National Natural Science Foundation of China under Grant 62073274; in part by the Guangdong Basic and Applied Basic Research Foundation under Grant 2023B1515020089; and in part by the Shenzhen Institute of Artificial Intelligence and Robotics for Society under Grant AC01202101103. (\textit{Corresponding author: Tin Lun Lam.}) }
\thanks{Guanqi Liang, Di Wu, Yuxiao Tu and Tin Lun Lam are with the School of Science and Engineering, The Chinese University of Hong Kong, Shenzhen, Guangdong, 518172; Shenzhen Institute of Artificial Intelligence and Robotics for Society (AIRS), Shenzhen, Guangdong, 518172, P.R. China (email:guanqiliang@link.cuhk.edu.cn; diwu1@link.cuhk.edu.cn; yuxiaotu@link.cuhk.edu.cn; tllam@cuhk.edu.cn).}
}

\markboth{ }
{Shell \MakeLowercase{\textit{et al.}}: Bare Demo of IEEEtran.cls for IEEE Journals}

\maketitle

\begin{abstract}
The intrinsic modularity and reconfigurability of modular reconfigurable robots (MRR) confer advantages such as versatility, fault tolerance, and economic efficacy, thereby showcasing considerable potential across diverse applications. The continuous evolution of the technology landscape and the emergence of diverse conceptual designs have generated multiple MRR categories, each described by its respective morphology or capability characteristics, leading to some ambiguity in the taxonomy. This paper conducts a comprehensive survey encompassing the entirety of MRR hardware and design, spanning from the inception in 1985 to 2023. This paper introduces an innovative, unified conceptual framework for understanding MRR hardware, which encompasses three pivotal elements: connectors, actuators, and homogeneity. Through the utilization of this trilateral framework, this paper provide an intuitive understanding of the diverse spectrum of MRR hardware iterations while systematically deciphering and classifying the entire range, offering a more structured perspective. This survey elucidates the fundamental attributes characterizing MRRs and their compositional aspects, providinig insights into their design, technology, functionality, and categorization. Augmented by the proposed trilateral framework, this paper also elaborates on the trajectory of evolution, prevailing trends, principal challenges, and potential prospects within the field of MRRs.
\end{abstract}


\IEEEpeerreviewmaketitle

\section{Introduction}\label{sec:intro}

\IEEEPARstart{M}{odular} Reconfigurable Robot (MRR) represents a distinctive class of robotic systems comprising multiple independent modules that can be reconfigured into various shapes by altering the interconnections among modules, thereby adapting to diverse tasks and environments\cite{yim2007modular}. Notably, MRR distinguishes itself from general multi-robot systems, where individual robots collaborate but remain physically disconnected\cite{rubenstein2014programmable}. In contrast, MRR modules can be physically interconnected, granting the system the ability to adopt numerous configurations, resulting in heightened adaptability and flexibility in task-solving\cite{seo2019modular}. Diverging from merely modular robotic systems\cite{ur5}, which integrate modular components primarily for simplified assembly and maintenance, MRRs encompass the unique capacity to undergo structural and functional reconfiguration\cite{daudelin2018integrated}, a trait responsive to evolving conditions or objectives.

\begin{figure*}
  \centering
  \includegraphics[width=\linewidth]{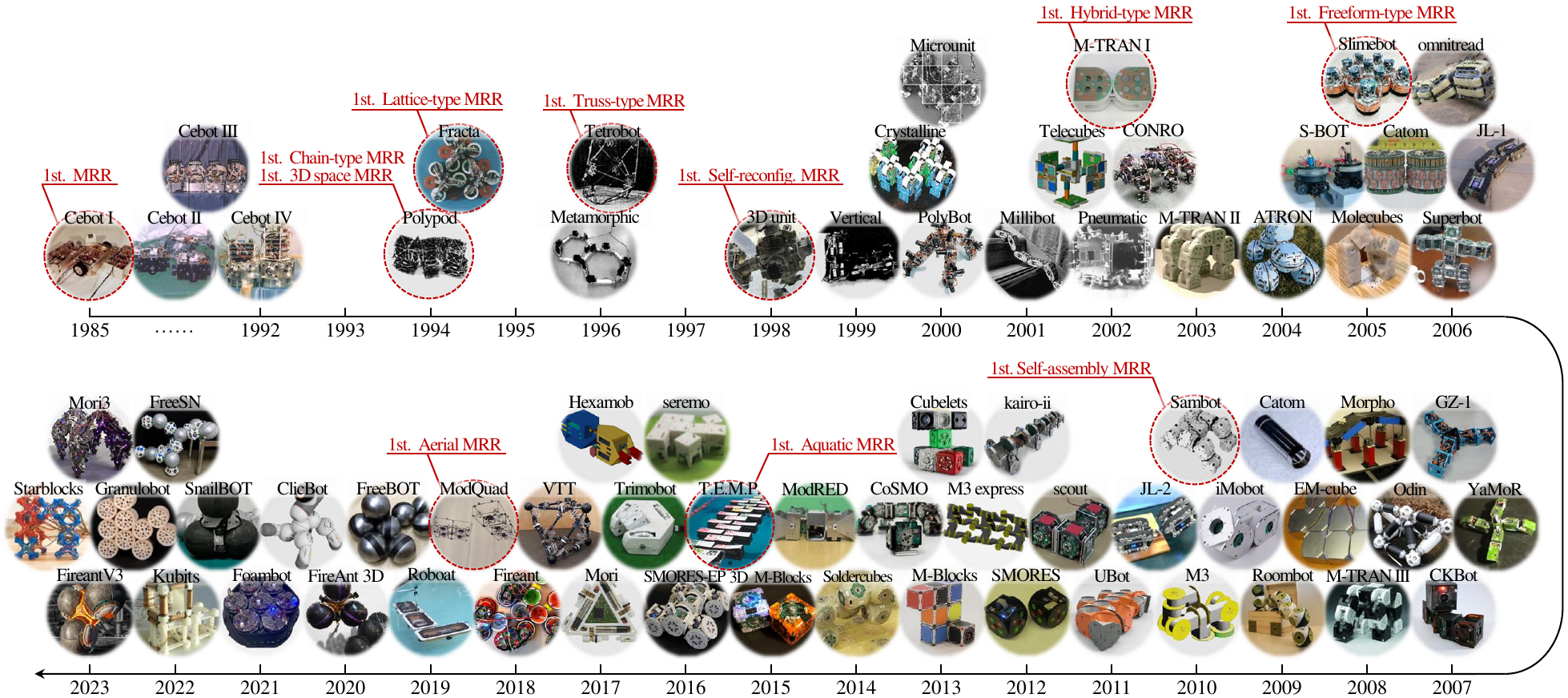}
  \caption{The chronicle of MRR spans from 1985 to 2023. The figure presents a chronological sequence of representative MRRs, enabling a visual assessment of their advancements and modifications over the years. Furthermore, it highlights significant achievements reached during this progression, emphasizing pivotal milestones.}
  \label{fig:timeline}
\end{figure*}

\begin{figure}
  \centering
  \includegraphics[width=.95\linewidth]{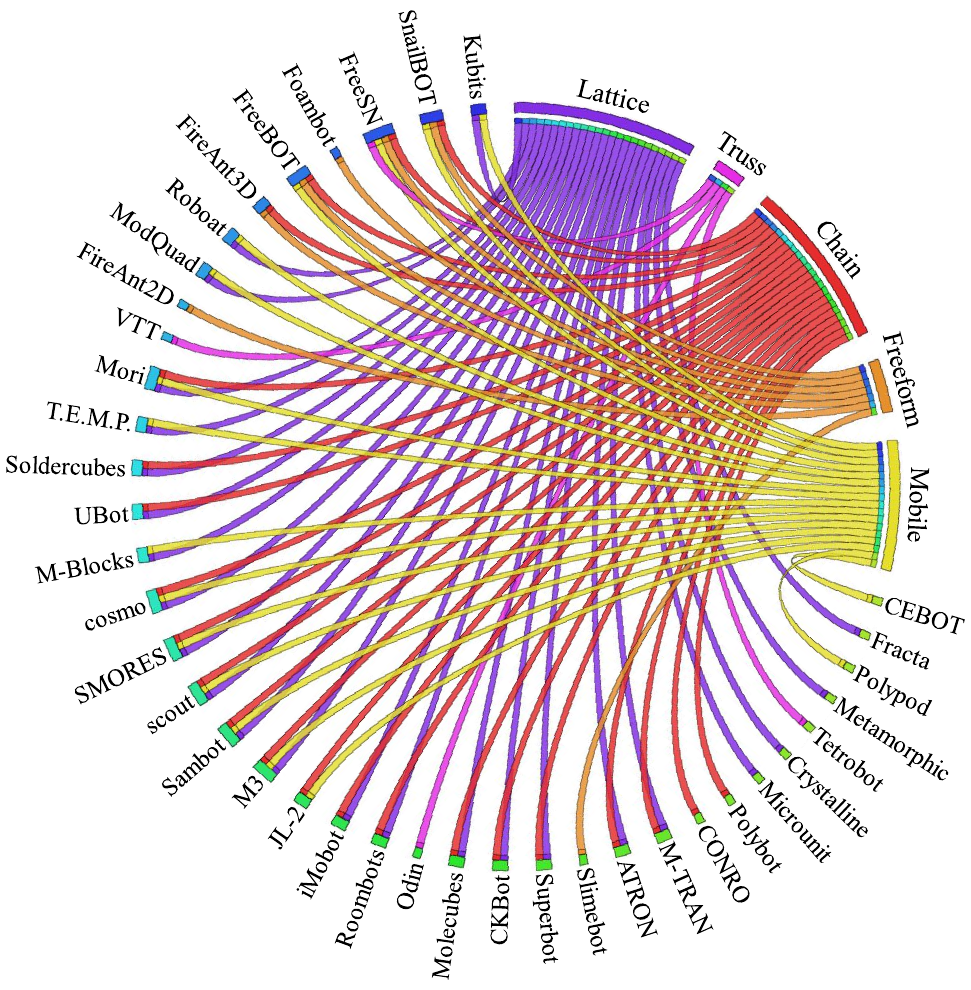}
  \caption{Affiliation of representative MRRs with documented taxonomic classifications. The presence of ambiguous and overlapping divisions is apparent in these classifications, as numerous MRRs simultaneously belong to multiple categories.}
  \label{fig:taxanomy}
\end{figure}


The reconfigurability of MRRs grants them diverse functionalities. MRR excels in challenging terrain exploration\cite{yim1995locomotion}, adept at negotiating complex environments, overcoming obstacles, and accessing hard-to-reach areas such as uneven ground\cite{swissler2020fireant3d}, ravines\cite{tu2022freesn}, stairs\cite{luo2020obstacle}, and waterways\cite{paulos2015automated}. 
Additionally, MRRs can adjust its shape for various object manipulation tasks, such as obstacle avoidance\cite{zong2022kinematics}, handling\cite{sugihara2023design}, grasping\cite{zhao2022snailbot}, and transportation\cite{tu2022freesn}. Another notable application of MRRs is in the realization of programmable matter\cite{piranda2018designing, zykov2005self}, where these versatile robotic units can be assembled into structures or objects of diverse shapes and functions. This finds practical application in areas such as reconfigurable furniture\cite{sprowitz2014roombots,hauser2020roombots}, addressing changing furniture needs and space constraints. Moreover, MRRs serve as programmable modular educational robots\cite{cubelets}, offering students a hands-on learning platform to explore robotics, programming, and engineering concepts. The future outlook for MRR is promising, with exciting possibilities such as modular building construction\cite{werfel2008three,lawson2014design}, exploration of outer space\cite{belke2023morphological,zykov2007evolved}, and the development of universal robots\cite{daudelin2018integrated}. Scientists have envisioned a groundbreaking concept of ``Bucket of Stuff''\cite{yim2007modular} , where MRR acts as highly adaptable and potent general-purpose robots, capable of transforming for diverse tasks across various fields.

Fig. \ref{fig:timeline} offers a comprehensive overview of the evolutionary progression of MRR over a span of nearly four decades. In 1985, the first MRR CEBOT\cite{fukuda1985cell,fukuda1988approach,fukuda1990cellular} emerged, featuring multiple units, each capable of independent approach, connection, and separation, pioneering the field of MRR. The captivating features of MRR have garnered significant interest, and as diverse designs and classifications continue to expand, the MRR landscape experienced continuous transformation. In 1994, Polypod\cite{yim1994new} achieved configuration diversification and functionalization through joint design, marking a groundbreaking milestone as the first MRR to operate in three-dimensional space and pioneering the concept of chain-type MRR. Concurrently, within the same year, Fracta\cite{murata1994self} has been introduced, recognized as the pioneering lattice-type MRR. In 1996, Tetrobot\cite{hamlin1996tetrobot} demonstrated variable truss using a combination of node-link elements, marking a significant milestone as the first truss-type MRR. In 1998, the 3D unit\cite{murata19983} showcased the capacity for self-reconstruction within three-dimensional space, marking yet another significant advancement in the field. In 2002, the M-TRAN series\cite{murata2002m,kurokawa2003m,kurokawa2008distributed} embodied both the characteristics of chain and lattice-type MRR, thereby pioneering the concept of the first hybrid-type MRR. In 2005, Slimebot\cite{shimizu2005slimebot} achieved the capability of unrestricted module connection in any direction through a fully enclosed Velcro connector design, heralding the emergence of the first freeform-type MRR. In 2010, Sambot\cite{wei2010sambot} showcased the self-assembly capabilities within MRR, employing a fusion of wheel drive and controllable connectors, signifying another significant milestone in the field. In 2015, Yim et al.\cite{paulos2015automated}  extended the workspace of MRR into aquatic environments, while in 2018, ModQuad\cite{saldana2018modquad} further expanded the workspace into aerial domains. Furthermore, the progression of algorithms\cite{yoshida2002self,park2008automatic,butler2003distributed} serves to augment and propel the advancement of hardware technologies. These contributions have greatly advanced the field of MRR, leading to interdisciplinary interest and a growing number of engaged research collectives in this domain\cite{bhattacharjee2021magnetically,li2019particle,zykov2005self}. 

However, the continuous advancement of MRR hardware, accompanied by the proliferation of diverse design paradigms, has introduced a degree of ambiguity within the established standards and classification frameworks within this domain. Fig. \ref{fig:taxanomy} provides an overview of affiliations between representative MRRs and taxonomic classifications as documented in the literature. 
These classifications encompass five widely recognized categories\cite{seo2019modular}: lattice, chain, mobile, truss, and freeform, each described by its respective morphology and capability characteristics.
Evidently, these categorizations exhibit perceptible vagueness and intersections, where a substantial cohort of MRRs belonging to multiple distinct categories. These categories of MRR depends on various criteria and is significantly influenced by the historical era. This is mainly because a specific pioneering MRR achieved remarkable performance during a particular period, shaping the prevailing design trends of that era and leading to the emergence of new categories. During the early phases of history, the categorization of MRR predominantly relied on morphological characteristics, categorizing it into chain\cite{yim1994new,yim2000polybot,jorgensen2004modular,davey2012emulating,murata2002m} and lattice\cite{murata1994self,pamecha1996design,yoshida2000micro,saldana2018modquad,cubelets} types, encompassing the predominant MRR types of that period.
At a subsequent phase, owing to the pronounced facet of independent mobility inherent to certain MRRs, a taxonomy centered around mobility-oriented attributes materialized\cite{fukuda1990cellular,brown2002millibot,mondada2004swarm,davey2012emulating,saldana2018modquad}.
Moreover, with the growing prevalence of MRRs featuring truss-like structural designs, truss morphology\cite{lyder2008mechanical,spinos2017variable,tu2022freesn,yu2008morpho,hamlin1996tetrobot} has emerged as a new classification aspect.
Later iterations saw the emergence of the freeform classifications 
\cite{liang2020freebot,swissler2020fireant3d,swissler2023fireantv3,tu2022freesn,shimizu2005slimebot}, due to the popularity of specific MRR connectors capable of connecting in various orientations and positions.
It's worth noting that certain MRRs, possessing characteristics spanning multiple classification domains, are labeled as hybrid-type MRRs.
Amidst the considerable advancements made by MRR in recent times through a profusion of innovative designs, a challenge endures in reconciling classification frameworks with established standards due to the ongoing evolution of technological paradigms and morphologies, thereby giving rise to new categorizations.
There is an urgent need for a concerted effort to establish a rigorously scientific and standardized taxonomy that aligns with the current state of technological advancement, addressing the congruence and criteria of robot design within this domain.

This paper strives to expound upon the evolution of MRR by surveying, assessing, classifying, and summarizing existing MRR hardwares. 
The primary objective is to develop an enhanced and systematically structured framework with the aim of providing practitioners, especially those new to the field, an additional technical perspective to better understand various MRR hardware. In parallel, the intention is to provide valuable guidance that streamlines research endeavors, fosters collaborative engagement among researchers, and expedites technological breakthroughs within the realm of MRR. In totality, this paper aims to address the following inquiries:

\begin{itemize}
  \item What are the commonalities and design principles among different MRR mechanisms and designs, and how can we view them with a unified perspective?
  \item How can these MRRs be classified more scientifically and standardized?
  \item What have been the past development trends and changes in hotspots for MRR, and what aspects should MRR strive for in the future?
\end{itemize}
Keeping the three aforementioned questions in consideration,  a survey of pertinent literature spanning the period from 1985 to 2023  (up to the present juncture) was undertaken. This paper presents a holistic survey of MRR hardware, encompassing its design, technology, capabilities, and even its taxadonomy. The principal focus is on decoding MRR to clarify their fundamental topological characteristics, as well as their constituent elements, alongside the characteristics stemming from their permutations and combinations. This categorization and differentiation of MRRs, based on their essential characteristics, aims to address the lack of standardization in the field. It seeks to enhance our understanding of how different types of MRRs relate to each other and what makes them unique, going beyond a simple categorization based on morphology and capability. Furthermore, in tandem with the outlined foundational elements, we conduct a thorough historical analysis of MRR, culminating in the synthesis of overarching developmental trajectories, functional shifts, and usage trends, thus highlighting future research prospects and persistent challenges in the field. By delving into existing works, we aspire to furnish readers with a thorough understanding of the multidimensional landscape characterizing the realm of MRRs. Succinctly stated, our main contributions are encapsulated in the ensuing points:
\begin{enumerate}
  \item We propose a novel and unified structured view of MRR hardware, decoding MRR into three basic elements: connectors, actuators, and homogeneity, and classifying them individually. This perspective allows us to provide clear definitions for some previously ambiguous terms and clarify the relationship between MRR and its categories in the past. It contributes to a more scientific and standardized classification of these robots.
  \item We offer an extensive overview of MRR hardware, encompassing recent advancements. By tracing the field's 40-year evolution in conjunction with outlined foundational elements, we pinpoint trends in design, function, and usage technologies, providing guidance for MRR designs and fostering enhanced MRR technology development.
  \item We outline several open questions and promising future research directions.
\end{enumerate}

The subsequent sections of this paper are structured as follows: In Section \ref{sec:Essential Elements of MRR}, three fundamental elements of MRR were put forth, along with the introduction of a novel tripartite conceptual framework. Subsequently, an disscussion of the interconnections between these elements and their alignment with previous taxonomic categorizations is conducted. Building upon the proposed tripartite framework, Section \ref{sec:tech2cap}  investigates the historical evolution of MRR, including technologies and advancements in capabilities. Section  \ref{sec:summary} furnishes a paper conclusion while also presenting potential research directions and associated challenges in the field of MRR.

\section{Decoding MRR: Tripartite Elements}\label{sec:Essential Elements of MRR}
Within this section, we present a novel conceptual framework concerning MRR, specifically decoding MRR to emphasize three key elements: connectors, actuators, and homogeneity. The role of connectors is  to integrate independent robotic modules into a unified whole, while actuators induce relative motion among these modules. The homogeneity aspect assesses the inherent differences in modules composing the system, thereby distinguishing whether it is constructed from a single module category or multiple module categories. 
The evolving MRR taxonomy, influenced by technological advancements, experienced some confusion.
However, the three basic elements proposed above emphasize intrinsic properties and provide a supplementary technical perspective, which aids in better understanding MRR.
This section provides a detailed explanation of these conceptual structures, fortified by illustrative examples of various MRR systems embodying these properties. Additionally, we assess the alignment between established, widely recognized MRR taxonomies and the new conceptual framework we introduce, thereby elucidating the interrelationships. By emphasizing fundamental properties, we aim to establish a more cogent foundational design principle for MRR, offering practical guidance for future MRR hardware development.

\begin{figure}
  \centering
  \includegraphics[width=\linewidth]{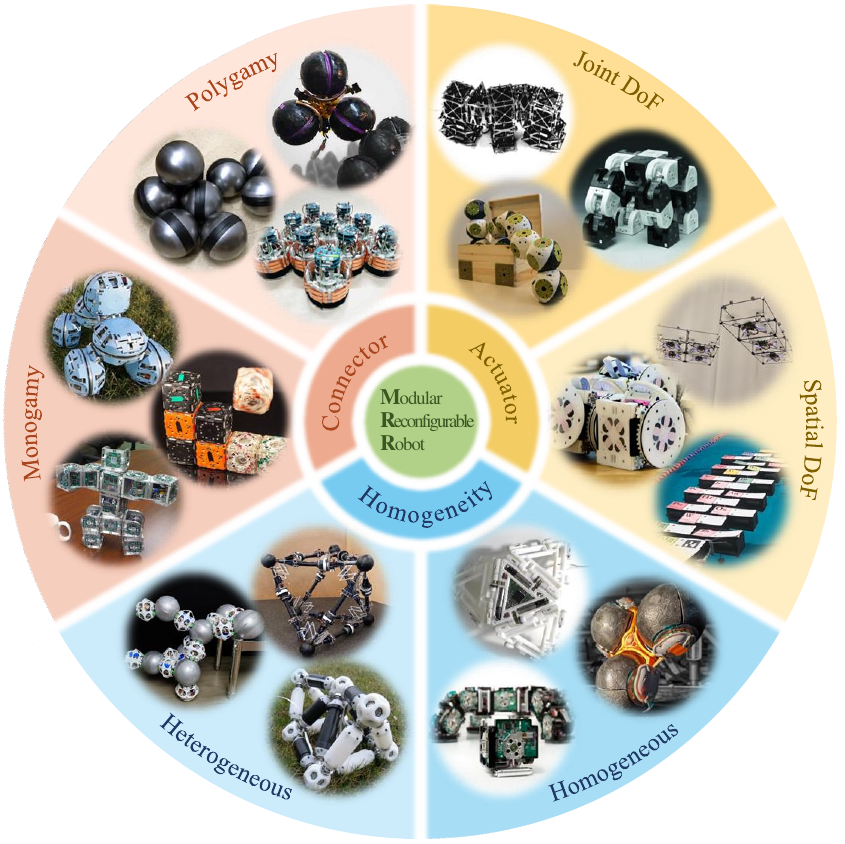}
  \caption{A novel tripartite framework decodes MRR systems into three elements: connectors,  actuators, and homogeneity. The classification of these three elements differs at various levels: connectors are classified based on the degree of the connection topology they form, actuators are classified by the type of DoF they provide, and homogeneity pertains to the system's classification based on its constitution, determining whether it contains a single type or multiple types of modules. Representative MRRs are provided for each element attribute.}
\end{figure}

\begin{figure*}
  \centering
  \includegraphics[width=.85\linewidth]{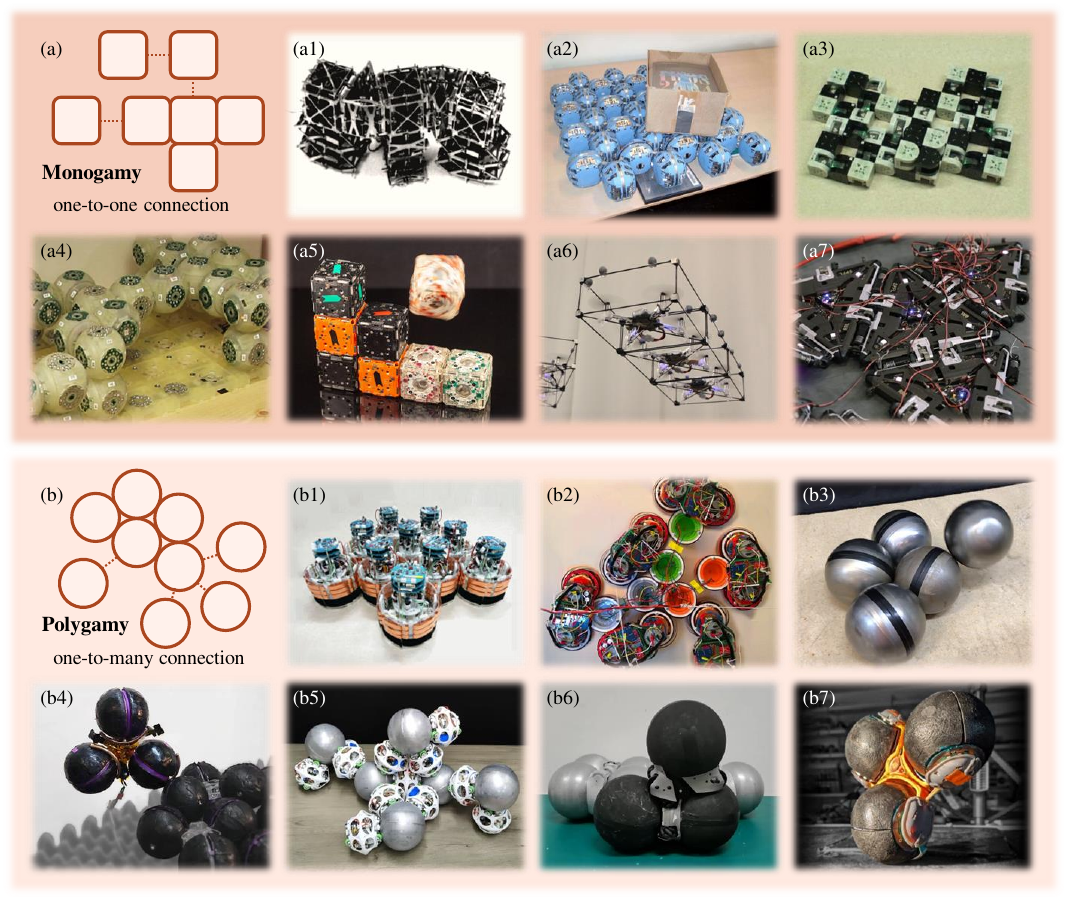}
  \caption{Connector elements are categorized as either (a) monogamy or (b) polygamy. Monogamous connectors establish one-to-one relationships among connectors. Noteworthy exemplars encompass: (a1) Polypod\cite{yim1994new}, (a2) ATRON\cite{jorgensen2004modular}, (a3) M-TRAN\cite{murata2002m} , (a4) Soldercubes\cite{neubert2014self}, (a5) M-Blocks\cite{romanishin2013m}, (a6) ModQuad\cite{saldana2018modquad}, and (a7) Mori3\cite{belke2023morphological}. In contrast, polygamous connectors create one-to-many relationships among connectors. Illustrative instances encompass: (b1) Slimebot\cite{shimizu2005slimebot}, (b2) FireaAnt2D\cite{swissler2018fireant}, (b3) FreeBOT\cite{liang2020freebot}, (b4) FireaAnt3D\cite{swissler2020fireant3d}, (b5) FreeSN\cite{tu2022freesn}, (b6) SnailBOT\cite{zhao2022snailbot}, and (b7) FireaAntV3\cite{swissler2023fireantv3}.}
\end{figure*}

\subsection{Connector}
Within the sphere of MRR, connectors fulfill the role of physically connecting modules. The inter-module topology is significantly shaped by the connector mechanism, emphasizing the pivotal importance of the properties and distribution of these connectors in MRR design. Previous surveys\cite{chennareddy2017modular,saab2019review} have examined various aspects of connectors, encompassing attributes such as gender, quantity, and tolerances, primarily concentrating on diverse docking mechanisms, evolving in parallel with technological advancements. Here, with reference to marriage customs in the field of cultural anthropology, we introduce novel  connector conceptualizations of monogamy and polygamy. Certain connectors facilitate exclusive one-to-one connections, whereas others have the capability to connect with multiple connectors simultaneously. This inherent characteristic of connectors directly impacts the structural topology and connection relationship within MRR systems, thereby justifying its inclusion as a fundamental element property in our analytical framework.

\subsubsection{\textbf{Monogamous Connector}}
The monogamy concept in connector design emphasizes the establishment of one-to-one connections between connectors. Connectors capable of forming a single connection with their corresponding counterparts are classified as monogamous connectors. This concept was widely adopted in the early stages of modular robot development and continues to be prevalent in most modern MRR systems.
For instance, the SMORES system\cite{davey2012emulating} consists of homogeneous modules equipped with four gender-less docking ports that utilize docking keys (later upgraded to electromagnets in SMORES-EP\cite{tosun2016design} ) as their connection mechanism, allowing for a face-to-face connection between the docking ports of two modules.
The Molecube system\cite{zykov2005self} consists of cubic modules, each equipped with six faces featuring an identical mechanical connector that utilizes pins and sockets for mechanical fixation, facilitating a face-to-face interconnection between modules.
The Polypod\cite{yim1994new} and PolyBot\cite{yim2000polybot} systems each consist of two module types: segment modules with two plates, and node modules with six plates. These plates enable a genderless connector that facilitates a face-to-face mechanical module connection.
The M-Blocks system\cite{romanishin2013m,romanishin20153d} utilizes cube modules equipped with cylindrical bonded magnets on all 12 edges, enabling face-to-face alignment and magnetic connections between neighboring modules.
ModQuad\cite{saldana2018modquad}, encompassing multi-rotors within cube-shaped docking frames, utilize magnets positioned at the frames corners to achieve face-to-face magnetic connections between the cube modules.
Mori\cite{belke2017mori, belke2023morphological} employs triangular modules equipped with a pin and socket docking mechanism on three sides, enabling the gender-less assembly of these triangular units.
While numerous historical examples of gender-neutral connectors exist, it's essential to emphasize that monogamy exclusively pertains to one-to-one connector properties, covering connections between connectors of different genders and those of the same gender. Specific monogamous connectors incorporate male and female components to facilitate module connections through the mating of opposite-gender connectors.
Illustratively, the UBot system\cite{zhao2011new} encompasses active and passive connectors; the former features motor-driven hooks, and the latter bears four slots, facilitating a gendered face-to-face mechanical connection via hook-slot pairings.
The Odin system\cite{lyder2008mechanical} consists of joint modules with ball-and-socket joints and six encircling connection slots, along with link modules featuring plug ends that align with these slots, enabling gendered face-to-face mechanical connections.

Monogamy or polygamy is an inherent property of the connector itself, independent of the module it belongs to. While certain modules may have multiple monogamous connectors to establish connections with multiple modules\cite{davey2012emulating,sproewitz2009roombots,romanishin2013m}, the connection characteristics of these modules does not involve our discussion of monogamous or polygamous connector properties. Connector attributes have a significant impact on the resulting MRR topology, a relationship often analyzed using graph theory. In graph theory representations, these connector properties can be classified as follows: when MRR connectors are gender-less, they lead to an undirected graph, while gendered connectors result in a directed graph. Additionally, a module equipped with n monogamous connectors will generate a topology with at most n edges, with gendered monogamous connectors forming directed edges and non-gendered monogamous connectors forming undirected edges. This distinction is of paramount importance when planning MRR system configurations and connection relationships\cite{luo2022auto}.

\subsubsection{\textbf{Polygamous Connector}}
In contrast to some previous approaches focused on augmenting MRR module connectivity by adding multiple connectors within one module, alternative research\cite{shimizu2005slimebot,swissler2020fireant3d,liang2020freebot,spinos2017variable} introduces a distinct paradigm where a single connector forms connections with multiple counterparts. This unique connector type highlights a one-to-many relationship among connectors, and we define connectors with this characteristic as polygamous connectors.
For example, Slimebots\cite{shimizu2005slimebot} are circular modules enclosed by Velcro straps that interconnect when the modules are in proximity, enabling a single Slimebot to establish a one-to-many connection, accommodating up to six modules along the circumference of the Velcro.
With similar uniform circular modules and conductive plastic continuous docks, FireAnt2D\cite{swissler2018fireant} establishes connections through a surrounding docking contact, enabling one FireAnt2D to link with up to 6 identical counterparts.
Moreover, FireAnt2D expanded to FireAnt3D\cite{swissler2020fireant3d} and FireAntV3\cite{swissler2023fireantv3}, introducing a 3D continuous docking system with three docks per robot. These docks use Joule heat to fuse the outer conductive plastic layer, facilitating multiple one-to-many connections along the spherical dock.
The aforementioned MRRs feature genderless polygamous connectors, enabling one-to-many connections among connectors of the same type. However, not all polygamous connectors are gender-less; some polygamous connections occur between connectors of opposite genders.
For instance, consider FreeBOT\cite{liang2020freebot,lam2021self}, a spherical robot enclosed in a low-carbon steel shell with internal magnets. It has the capability to connect to another FreeBOT's entire shell through magnetic attraction. The steel shell can accommodate up to 11 FreeBOT magnets connecting to it, highlighting that the connection in FreeBOT is a gendered polygamous connection.
Snailbot\cite{zhao2022snailbot}, inspired by the snail's morphology, features magnets on its chassis and a rear-mounted low-carbon steel sphere. Using a magnet-sphere connection akin to FreeBOT, each Snailbot's sphere can connect to the chassis of up to three other Snailbots, resulting in a gendered polygamous connection.
In the FreeSN system\cite{tu2022freesn}, the struct module can establish a similar magnetic-sphere connection with the steel sphere node module, allowing for a capacity of up to 12 connections among them, thus creating a one-to-many gendered polygamous connection.
The VTT system\cite{spinos2017variable} consists of node modules and member modules. By utilizing a zipper mechanism, multiple members can be attached to a single node module on its spherical surface, thereby forming the gendered polygamous connection.
The one-to-many connection capability of the polygamous connector enables a robot module equipped with just one such connector to establish connections with multiple other modules.
It's important to emphasize that a single module can also accommodate multiple polygamous connectors. For example, both the FireAnt3D\cite{swissler2020fireant3d} and FireAntV3\cite{swissler2023fireantv3} modules feature three identical solder-based polygamous connectors, while the struct module in the FreeSN system\cite{tu2022freesn} includes two magnet-based polygamous connectors.
Regarding the topology aspect, it's evident that polygamous connectors facilitate the robot's ability to establish multi-degree topologies, with the maximum topological degree determined by the maximum number of connectable modules within the MRR system.
While such a multi-degree topology previously required multiple monogamous connectors within MRR, this can now be achieved through the implementation of a single polygamous connector.
In a manner akin to the earlier discussion on monogamous connectors, the gender of connectors impacts graph order: modules equipped with gender-specific polygamous connectors yield ordered graphs, whereas those with gender-less polygamous connectors lead to unordered ones.
For example, in both FreeBOT\cite{liang2020freebot,liang2022energy} and SnailBOT\cite{zhao2022snailbot} cases, the magnet-sphere connection allows one sphere to attach to multiple magnets, creating a gendered multi-module connection with a directed multi-degree topology\cite{luo2022auto}.
Furthermore, the characteristics of the polygamous connector share similarities with the freeform-type MRR, which will be further discussed in Section \ref{sec:discussion}.

\begin{figure*}[]
  \centering
  \includegraphics[width=0.85\linewidth]{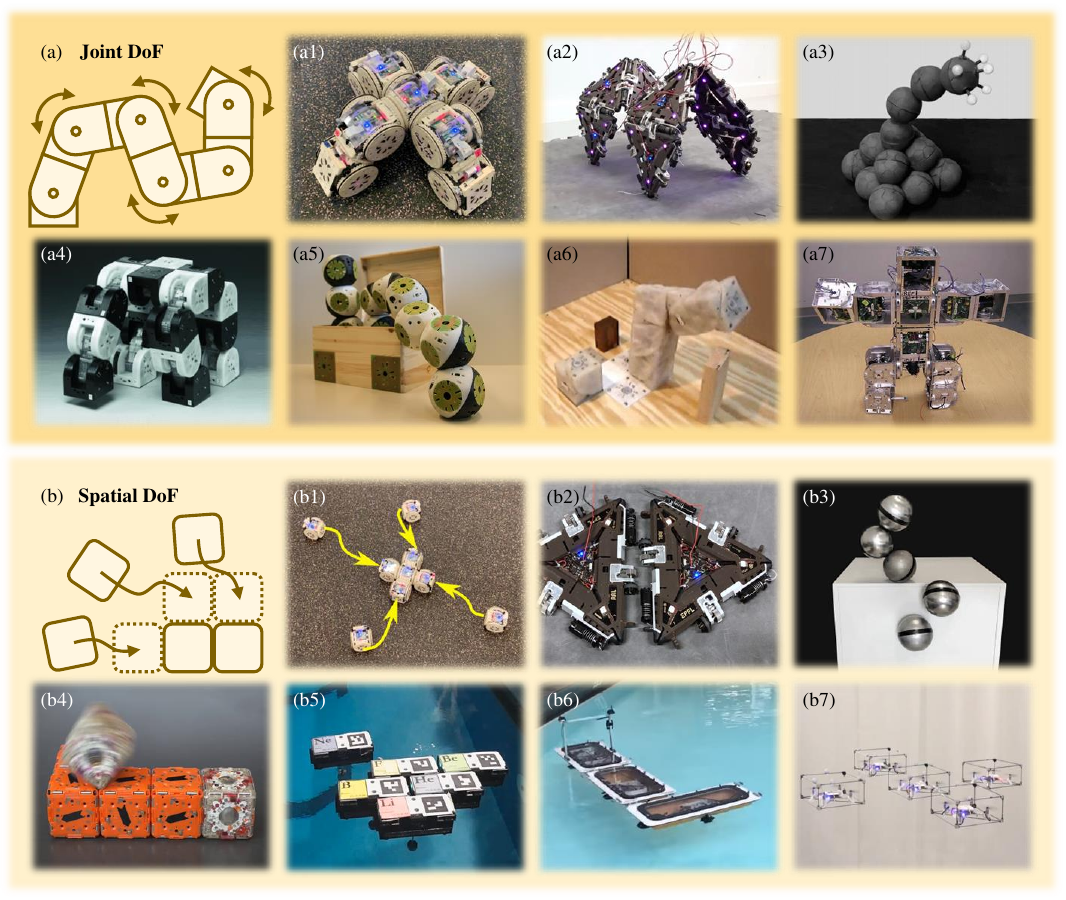}
  \caption{Actuator elements are recognized to possess two distinct types of DoF: (a) joint DoF and (b) spatial DoF. The joint DoF pertains to the relative extent of liberty demonstrated by connectors situated within a module. Noteworthy exemplifications include the following: (a4) M-TRAN \cite{murata2002m}, (a5) Roombots \cite{sproewitz2009roombots}, (a6) Molecubes \cite{zykov2005self}, and (a7) Superbot \cite{salemi2006superbot}. Conversely, the spatial DoF is concerned with the independent moving capacity of a MRR module within an expansive spatial domain. Eminent instances within this classification comprise: (b4) M-Blocks \cite{romanishin2013m}, (b5) T.E.M.P. \cite{paulos2015automated}, (b6) Roboat \cite{wang2019roboat}, and (b7) ModQuad \cite{saldana2018modquad}. It is worth mentioning that some MRR modules have both DoF: (a1)$\&$(b1) SMORES \cite{davey2012emulating}, (a2)$\&$(b2) Mori3 \cite{belke2023morphological}, and (a3)$\&$(b3) FreeBOT \cite{liang2020freebot}. }
\end{figure*}

\subsection{Actuator}
The motion of the MRR system, encompassing reconfiguration\cite{kurokawa2008distributed,white2010reliable} and locomotion\cite{fitch2008million,sproewitz2008learning,dumitrescu2004formations}, is achieved by adjusting the relative positions of robot modules to attain DoF, which are facilitated by the actuators.
In this section, we will offer a comprehensive explanation of the actuator within the proposed MRR system's tripartite framework, covering its concept, effects, and illustrative examples.
For the first time, we classify the role of actuators into two types, considering their impact on the DoF properties of the MRR system: spatial DoF and joint DoF. The DoF discussed here pertain to an individual MRR module, not the MRR system as a whole. When multiple modules come together to form an MSRR system, the combination of actuators results in multiple DoF as well.
The DoF in all historical MRR systems can be summarized as either spatial DoF\cite{fukuda1990cellular,romanishin2013m,saldana2018modquad,paulos2015automated}, joint DoF\cite{yim1994new,yim2000polybot,castano2002conro,baca2014modred}, or a combination of both\cite{murata2002m,davey2012emulating,belke2017mori,zykov2005self}.
Notably, this dichotomy concerning DoF underscores the inherent characteristics of the two historically popular MRR classifications: chain-type\cite{saldana2018modquad,paulos2015automated,romanishin2013m,hauser2020kubits} and lattice-type\cite{yim1994new,yim2000polybot,castano2002conro,sugihara2023design}.
The emphasis on these two types of DoF underscores the quintessence of actuators in the context of MRR, while also providing a perspective for comprehending actuators and their impact on the MRR system's ability to locomote and reconfigure.

\subsubsection{\textbf{Joint DoF}}
The concept of joint DoF involves the joint motion between connectors within a module, enabling the adjustment of the relative position and orientation between connectors.
It can provide relative DoF between connectors within the module, permitting joint motions like rotation or translation. When these modules are interconnected via connectors, the actuator can then achieve relative movement DoF between modules.
This type of DoF is closely associated with chain-type MRRs. Chain-type MRRs are those in which modules can be connected to form a chain or branched morphology\cite{yim2002connecting,luo2022auto,hou2008distributed}. They are equipped with an actuator that facilitates this DoF, enabling the robot to attain chain motion.
MRRs offer joint DoF through various actuator designs and technologies. Most MRR systems are equipped with one or several specialized actuators to enable this type of degree of freedom.
For example, in the Polypod system\cite{yim1994new}, the segment module consists of two semi-cylindrical components connected by a DC motor, which enables the components to rotate coaxially, forming a revolute joint.
M-TRAN have evolved by enhancing connectors over iterations\cite{murata2002m,kurokawa2003m,kurokawa2008distributed}, while retaining a similar core mechanism: two semi-cylindrical boxes, three connectors, and two servo motors for independent box rotation, which facilitates joint movement between connectors.
The ATRON\cite{jorgensen2004modular}, which is approximately spherical in shape, incorporates a central DC motor for rotating its hemispheres around a perpendicular axis at the equator. Each hemisphere accommodates four claw connectors, and the motor facilitates coaxial joint movement between these connector groups.
SuperBot\cite{salemi2006superbot} consists of two square blocks, with each block housing 3 connectors and 3 DC motors capable of achieving multi-DoF relative rotation.
Roombot\cite{sproewitz2009roombots} is also composed of two squares and is equipped with two diagonal DC motors, enabling relative rotation between three groups of ten connectors in total.
Molecube\cite{zykov2005self} have a lattice-shaped morphology and are divided into two diagonal triangular pyramidal halves. A centrally located servo motor enables relative rotation between the two halves around a shared axis.
FireAnt3D\cite{swissler2020fireant3d}features three solder-based sphere connectors and is actuated relative to the base and arm by a DC motor, offering rotational DoF among the three connectors.
Apart from the independent actuator inherent in enabling joint DoF, there are instances where, upon module connection, certain actuators originally serving different functions contribute to these such DoF, such as:
FreeBOT's\cite{liang2020freebot} internal driving mechanism enables it to roll independently, and when the modules are connected, it propels itself to roll on another FreeBOT, achieving relative rolling contact joint motion\cite{zong2022kinematics}.  In this context, the sphere and the magnet of can be viewed as two distinct connectors, and the joint motion provides joint DoF between them.
Likewise, the skeleton of StarBlocks\cite{zhao2023starblocks} is elastic and will deform and expand when electrically heated. As a result, the modules can squirm independently, and when combined, they can also achieve revolute joint movements, providing joint DoF between connectors.
Utilizing MRRs equipped with joint DoF actuators can enhance task processing capabilities, enabling various actions like manipulation\cite{zong2022kinematics,sugihara2023design}, transportation\cite{tu2022freesn}, and locomotion\cite{yim1994new,shokri2015meta}. We will introduce the actuator technologies related to joint DoF in more detail in Section \ref{sec:act tech}.

\subsubsection{\textbf{Spatial DoF}}
Spatial DoF in the MRR system enables the module to have independent DoFs for movement, allowing it to change both its position and attitude in space.
By moving independently in space, these modules can alter their spatial distribution and connection relationships, thus reconfiguring the system. 
MRR possessing spatial DoF resemble a particular type of mobile multi-robot systems, inheriting advantages such as scalability, flexibility, and adaptability to operate in expansive and dynamic environments\cite{zhou2022swarm}. Modules can adjust their positions for configurations, allowing tailored functions to align with mission goals and enhancing reconfigurability.
Various actuation solutions were employed to provide MRRs with spatial DoF, and many MRRs have showcased independent mobility in 2D space.
CEBOT\cite{fukuda1990cellular}, the earliest MRR, employs a differential wheel drive to provide spatial DoF and relative position adjustment in a 2D plane, thus facilitating system reconfiguration.
SMORES\cite{davey2012emulating} features four gender-less monogamous connectors, with two of them taking on circular shapes distributed within differential wheels and equipped with friction rubber rings on the edges. These can function as wheels, granting spatial DoF in a 2D plane, and enabling system self-assembly and self-configuration.
Sambot\cite{wei2010sambot} has two DC motor-driven wheels at its base, providing spatial DoF in a 2D plane. When combined with the four monogamous connectors it is equipped with, it can achieve self-assembly among modules.
Mori's\cite{belke2017mori} triangle is equipped with stepper motors and friction wheels on each of its three sides, and their coordination allows Mori to move omnidirectionally within the plane.
M-Blocks\cite{romanishin2013m} are cubic modules equipped with reaction wheels driven by motors to generate torque pulses, enabling them to achieve rolling motion on various surfaces, including the ground and other M-Blocks. This design provides spatial DoF for movement and reconfiguration.
Similarly, driven by an internal differential drive mechanism, FreeBOT\cite{liang2020freebot} can traverse various surfaces, including the ground, ferromagnetic slopes or walls, and even other FreeBOT surfaces, thereby providing functional spatial DoF for movement.
Furthermore, the spatial DoF in the MRR are also applicable in alternative environments, such as on water surfaces.
T.E.M.P\cite{paulos2015automated} is a square surface vessel designed to operate on water, with motors and impellers installed at the corners. Their coordination enables the modules to move and rotate independently, offering spatial DoF on the water and resulting in various system distributions and configurations.
Roboat\cite{wang2019roboat} is another module designed to operate on water, driven by a propeller, and its prowess in perception and planning further amplifies the contribution of spatial DoF on the water.
Moreover, certain MRRs have extended the spatial DoF into 3D space, demonstrating their capability for aerial locomotion.
ModQuad\cite{saldana2018modquad} is a quadcopter enclosed by a magnetic cube frame, powered by four motors, which offers spatial DoF in the air, facilitating dynamic system reconfiguration and good manipulation capabilities in 3D space.
The recent TRADY\cite{sugihara2023design} integrates magnetic connectors and controllable joints into the quadcopter, where the quadcopter offers spatial DoF in the air, and the joints provide joint DoF, demonstrating greater potential for powerful manipulation applications. 
Various technologies have been employed to provide spatial DoF for MRRs in specific environments, and a more detailed discussion of these actuator technologies will be explored in Section \ref{sec:act tech}. 
Furthermore, categorizing the spatial DoF as separate actuator classifications is in alignment with the typical characteristics of conventional mobile-type MRRs. We will explore their relationship in more detail in Section \ref{sec:discussion}.

\begin{figure*}
  \centering
  \includegraphics[width=.85\linewidth]{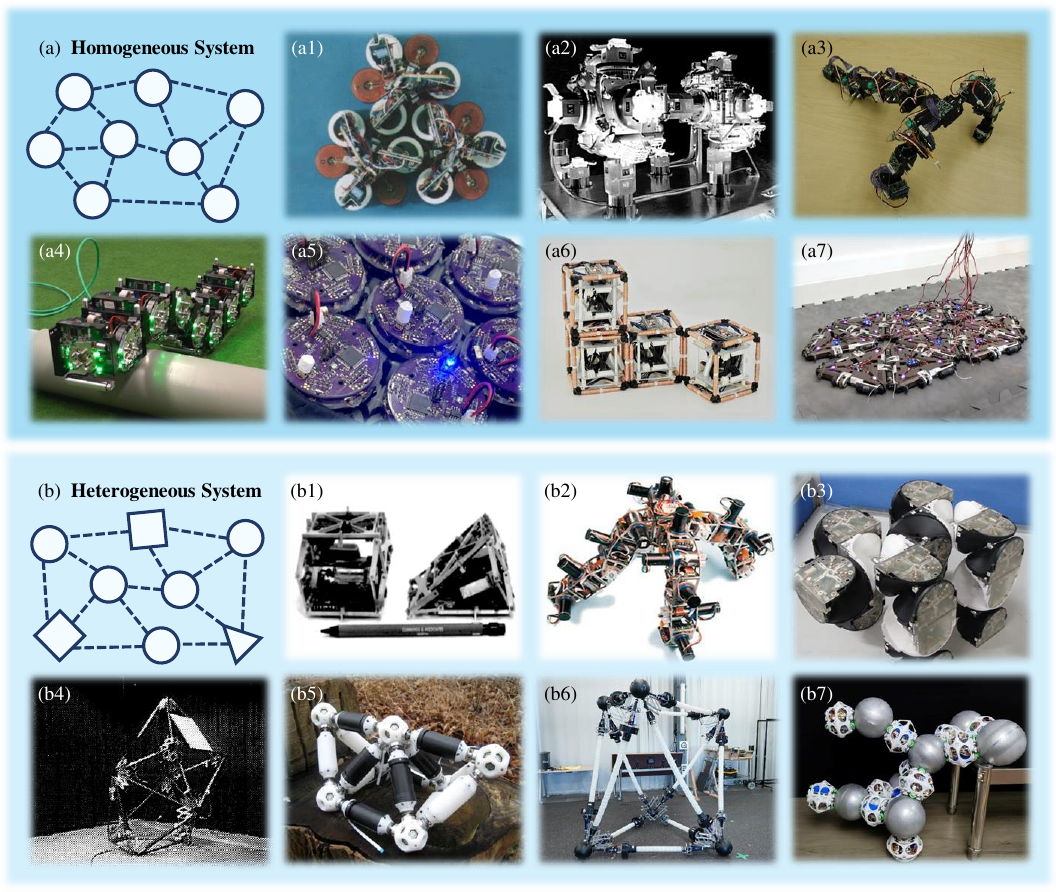}
  \caption{The taxonomy of homogeneity elements encompasses two main categories: (a) homogeneous  and (b) heterogeneous systems. Homogeneous systems indicate that they exclusively comprise one type of module. Illustrative instances include: (a1) Fracta\cite{murata1994self}, (a2) 3D unit\cite{murata19983}, (a3) CONRO\cite{castano2002conro}, (a4) CoSMO\cite{liedke2013collective} , (a5) Foambot\cite{ceron2021soft}, (a6) Kubits\cite{hauser2020kubits}, and (a7) Mori3\cite{belke2023morphological}. Conversely, heterogeneous systems utilize multiple types of modules. Exemplary manifestations encompass: (b1) Polypod\cite{yim1994new}, (b2) PolyBot\cite{yim2000polybot}, (b3) UBot\cite{zhao2011new}, (b4) Tetrobot\cite{hamlin1996tetrobot}, (b5) Odin\cite{lyder2008mechanical}, (b6) VTT\cite{spinos2017variable}, and (b7) FreeSN\cite{tu2022freesn}.}
\end{figure*}

\subsection{Homogenity}
The previous description of actuators and connectors outlined the essential characteristics of the miniature module within the MRR system. At the systemic level, whether the constituent modules are the same or different, will also have an impact on functions and properties.
We regard module homogeneity as a significant metric for MRRs and introduce it in this section.
MRR modules can be classified into two groups, either homogeneous or heterogeneous, depending on whether the miniature modules within the robotic system are uniform. These miniature modules are repeatable and compatible with connectors for interconnection. While historical MRRs were primarily homogeneous, recent years have witnessed an increasing prevalence of heterogeneous MRRs\cite{spinos2017variable,zhao2023meta,tu2022freesn}. Module homogeneity introduces fault tolerance and repeatability to the system, while module heterogeneity can enhance the system's functionality and performance. Both types of MRR are of interest, and their combined presence is expanding the MRR research community.

\subsubsection{\textbf{Homogeneous}}
Homogeneous MRRs consist of repeated identical miniature modules.
In the historical development of MRR, the majority of designs have been homogeneous\cite{murata2002m,sproewitz2009roombots,davey2012emulating,romanishin2013m,belke2017mori}, where all robot modules adhere to a unified mechanism design, encompassing connectors, actuators, and their arrangement.
It's worth noting that the built-in actuators of certain MRR modules can induce deformation in the module's shape when providing joint DoF.
As an illustration, For example, the soft modular robot Foambot\cite{ceron2021soft} is enveloped in a cylindrical, stretchable outer membrane, enabling it to expand and contract, which alters its effective elastic modulus and enhances compliance. This capability allows the robot swarm to move and demonstrates applications in filling through deformation.
StarBlocks\cite{zhao2023starblocks} is an alloy spring-pulled elastic star-shaped module connected via magnets at the corners. These springs deform and adapt when electrically heated, allowing the module to perform various movements, bends, and size adjustments.
The newly developed Mori3\cite{belke2023morphological} , building upon its predecessor\cite{belke2017mori} , introduces an additional actuator for single-body deformation, enhancing the module's capability for self-deformation and enabling a series of modular origami operations.
Although these MRR modules can alter their shape through internal actuators, enabling the deformation of the module itself, the identity of its fundamental elements maintains the module's homogeneous nature. Therefore, we regard the system composed of these robotic modules as remaining homogeneous, even in light of individual deformations of the modules.
Due to their uniform module design and composition, homogeneous MRRs demonstrate robust fault tolerance and redundancy. In the event of a module failure, the system can reconfigure itself, with any operational module stepping in to replace the faulty ones, thereby enhancing overall fault tolerance and system robustness.
Furthermore, homogeneous module design simplifies control and coordination\cite{luo2022auto,yao2019reconfiguration}, placing fewer limitations on system behaviors such as movement and reconfiguration. This also enhances adaptability across diverse tasks and environments with miniature reconfiguration requirements.

\subsubsection{\textbf{Heterogeneous}}
Heterogeneous MRRs encompass MRR systems that include multiple categories of minimal modules.
Early Polypod\cite{yim1994new} and Polybot\cite{yim2000polybot} consisted of two module types: segment and node. The segment modules, driven by DC motors, facilitated end-to-end dynamic 2-dof revolute joint motion. The nodes, cube-shaped and lacking actuators, served as power hubs. This differentiation in module roles clarified that segments were responsible for movement, nodes for power supply, and their combination enhanced the system's functionality and topology.
The contemporary truss-type MRR, which has gained popularity recently, can be regarded as a heterogeneous MRR system formed by two distinct minimal module types.
The VTT system\cite{spinos2017variable} consists of node modules and member modules. Nodes facilitate alignment between members to create ball joints, while members serve as structural elements with telescoping linear actuators. This differentiated design enables nodes to facilitate flexible truss configurations, while member modules adjust truss length and structure. Their collaboration allows for dynamic shape changes through node merging and splitting.
FreeSN\cite{tu2022freesn} comprises node and struct modules. Node module is a steel sphere with a built-in magnetic sensor array for configuration detection. Struct module consists of a pair of magnet connectors with chassis equipped with a two-wheel differential drive, allowing for ball joint movement on the surface of the node. This synergy results in adaptable, multifunctional robotic structures that can adjust to various tasks and changing environments.
Nonetheless, it's important to note that similarities in appearance don't indicate heterogeneity. Our emphasis is on the consistency of inherent properties that arise from the miniature module mechanisms.
Consider Ubot\cite{zhao2011new}, composed of visually similar yet functionally distinct active and passive modules, rendering it a heterogeneous MRR system. Both visually identical cube modules feature a motor-driven 1-DoF revolute joint and four connectors. The distinction lies in the active modules being equipped with motor-driven connectors, while the passive modules remain inert. Connection compatibility mandates that adjacent modules should be of opposite types.
It's notable that when various MRRs with distinct designs, but still compatible connectors, work together, they can also be seen as constituting a heterogeneous MRR system.
For instance, the FreeBOT\cite{liang2020freebot}, FreeSN\cite{tu2022freesn}, and SnailBOT\cite{zhao2022snailbot} robot series showcase distinct attributes and capabilities. For example, FreeBOT's rolling contact motion offers heightened dexterity, while FreeSN's truss structure enhances stability. However, these MRRs all adhere to the common connection principle of magnet and sphere, positioning them as a collaborative heterogeneous MRR system.
Another comparable example can be observed with Scout\cite{russo2012design} and CoSMO\cite{liedke2013collective} . Despite being distinct MRRs, they utilize connectors that are mutually compatible, enabling the interconnection of these two modules. Consequently, the amalgamation of Scout and CoSMO can be regarded as a heterogeneous MRR system.
In contrast to homogeneous MRR systems, heterogeneous module configurations impose planning limitations because of the additional connections and reconstruction constraints they introduce. Nevertheless, they enhance system functionality and performance by promoting specialization among modules, thereby fostering efficiency and high-performance attributes.

\subsection{Discussion}\label{sec:discussion}
Over the years, the landscape of MRRs has continually evolved, experiencing a constant influx of diverse designs and classifications.
As discussed in Section \ref{sec:intro}, the existing classifications of MRRs encompass lattice-type, chain-type, mobile-type, truss-type, freeform-type, and even hybrid-type. These classifications exhibit unclear boundaries between categories and significant overlap.
Within this section, we revisit previous popular MRR classifications using the newly proposed tripartite framework.
Our focus is on clarifying the distinctions among these prior classifications and their interconnections, especially in terms of their alignment with our proposed fundamental elements. In this context, we offer precise definitions for these conventional classifications and clarify the relationship between these traditional categories and our fundamental elements. The pertinent information is consolidated in Table \ref{tab:link}.
We hope that this reconclusion can offer an alternative perspective, aiding in a deeper technical comprehension of MRR, while also fostering advancement in the MRR field by elucidating the fundamental design principles and practical applications of these MRR types.

\subsubsection{\textbf{Lattice}}
Lattice-type MRR pertain to modules with a lattice morphology, where the strategic arrangement of connectors achieves a lattice-like combination of modules. This category represents an early stage in the development of MRRs.
Fracta\cite{murata1994self} stands as the first documented instance of lattice-type MRR in the literature.
Lattice-type MRR places its emphasis on interactions among adjacent modules and offers unique modeling and control methods. In terms of modeling, the workspace of lattice-type MRR is divided into discrete lattice regions\cite{butler2004generic}, simplifying system analysis. Concerning movement strategy, a key feature of Lattice-type MRR is the ability to achieve displacement capability through self-reconfiguration by altering the connection relationships\cite{fitch2008million,thalamy2019scaffold}.
The ability of lattice-type MRR systems to form lattice structures showcases their versatility and scalability in various configurations and morphologies.
Within the proposed tripartite framework, traditional lattice-type MRRs can be seen as featuring the characteristics of multiple monogamous connectors.
Notably, the FireAnt series MRR\cite{swissler2020fireant3d,swissler2023fireantv3} utilizes spherical connectors and lacks a lattice-like structure with multiple connectors. Therefore, the author classifies the FireAnt series as a non-lattice-type MRR.
Coincidentally, we hold a shared perspective: In accordance with our tripartite framework, the connectors of the FireAnt Series MRR are classified as polygamous connectors, which renders it inconsistent with our definition of a lattice-type MRR characterized by multiple monogamous connectors, leads us to the conclusion that it is not a lattice-type MRR.
Essentially, the classic lattice-type MRR is defined based on module morphology, whereas our approach strives to decode the ``lattice-like" feature from MRR elements. This reveals a connection between traditional lattice classification and our proposed monogamous connector type.

\subsubsection{\textbf{Chain}}
Chain-type MRR features modules with a string-like morphology, allowing for flexible configurations such as chains, branches, or snake-like arrangements. The actuators provide joint DoF to enable chain motion of the connected modules.
The morphology of chain-type MRRs also introduces distinctive modeling and control techniques. 
Regarding modeling, the integration of joint DoF from the individual modules embodies a multi-DoF system with versatile kinematics. Concerning control, the various connection configurations between modules correspond to multiple arrangements and movement strategies, such as quadrupeds\cite{whitman2023learning} , snakes\cite{wright2007design} , loop\cite{sastra2009dynamic} , and more, addressing diverse environments and needs.
In our proposed tripartite framework, chain-type MRRs incorporate actuators that offer joint DoF, facilitating relative motion between connectors within the module; these actuators also enable extended chain motion when the modules are interconnected.
Polypod\cite{yim1994new} marked the pioneering stage of documented chain-type MRRs, subsequently followed by a series of popular chain-type MRRs\cite{yim2000polybot,castano2002conro,brown2002millibot,mondada2004swarm,granosik2005omnitread,baca2014modred}, all characterized by the unique feature of joint DoF.
The recent FreeBOT\cite{liang2020freebot} and SnailBOT\cite{zhao2022snailbot} , despite lacking independent joint actuators, are still classified as chain-type MRRs because they can move on a peer surface and then form rolling contact\cite{zong2022kinematics} and spherical joint respectively, thereby offering joint DoF.
In summary, past chain-type MRR classification relied on chain-like properties, while our tripartite framework focuses on using motion DoF to decode chain-like features. These cases demonstrate strong consistency between traditional chain classification and our concept of joint DoF.

\begin{table*}
  \centering
  \caption{The connection between the tripartite framework and existing taxonomies.}
  \includegraphics[width=.9\linewidth]{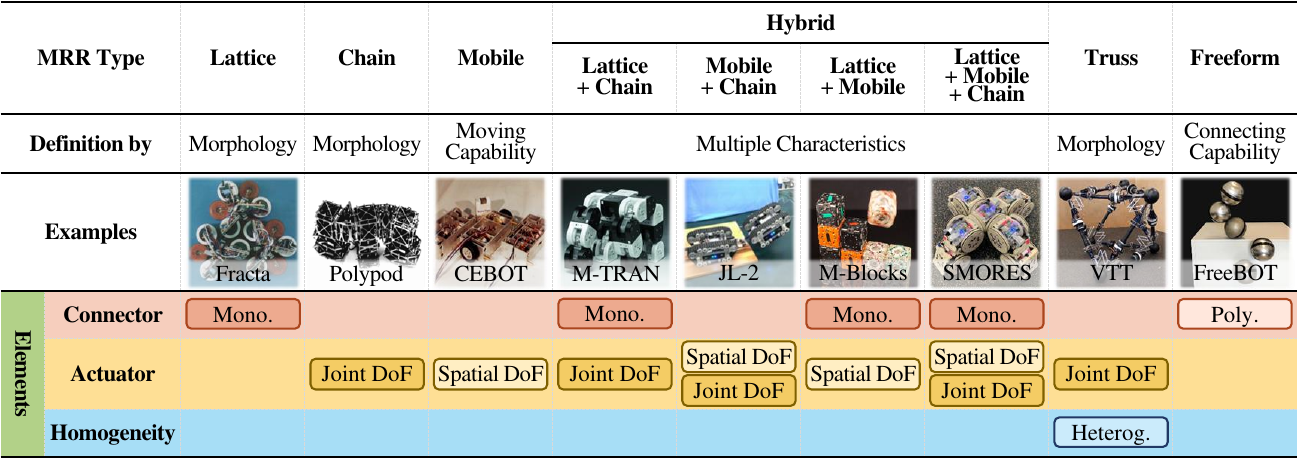}
  \label{tab:link}
\end{table*}

\subsubsection{\textbf{Mobile}}
Mobile-type MRR represents the convergence of MRR and traditional mobile multi-robots. It is characterized by modules within the system possessing the same moving capability as mobile robots, enabling them to move independently within the environment.
The increased mobility and adaptability inherent in mobile-type MRRs, operating as a distributed multi-robot system, present significant potential for achieving efficient task completion, including distributed collaboration, perception, monitoring, and more.
The capability of modules to move independently also facilitates self-assembly\cite{wei2010sambot,liu2023smores}, which is a key function of MRR. Modules can be connected and assembled from a non-connected state through their own mobility capabilities.
Within our proposed tripartite framework, mobile-type MRRs are equipped with actuators that provide spatial DoF, enabling each module to move independently in their surrounding environment.
The pioneer in the mobile-type MRR is CEBOT\cite{fukuda1990cellular}, initially designed as a physically connectable mobile multi-robot system. It also pioneered the MRR genre and represented a significant milestone in the field.
In the early days of the MRR field, the focus in terms of mechanism and design was primarily on the movement and reconfigurability of connected MRRs, with limited attention given to the mobility of individual modules. 
As the value of self-assembly capabilities became apparent, a series of MRRs with spatial DoF were introduced\cite{wei2010sambot,russo2012design,davey2012emulating,liedke2013collective,liang2020freebot,belke2023morphological}. These innovations showcased the remarkable applications of self-assembly capabilities within the realm of MRRs.
The focus on mobile-type MRR also reflects, to some extent, the significance of spatial DoF in MRR and the rationale for including it as one of the basic elements of the tripartite framework.

\subsubsection{\textbf{Hybrid}}
Hybrid-type MRR is a special category of MRR that simultaneously exhibits characteristics from various types mentioned above. Its primary feature is the composite fusion of two or even more types of characteristics, such as lattice-type, chain-type, and mobile-type.
As mentioned earlier, lattice-type, chain-type, and mobile-type MRRs each possess distinct advantages, while hybrid MRRs combine these characteristics to achieve scalability, mobility, and flexibility. 
In the past, these MRRs were defined based on various criteria, such as geometric similarity or specific definitional capabilities. Furthermore, throughout history, several types of hybrid MRRs existed, each representing hybrids between different types and displaying distinct characteristics. Our proposed tripartite framework not only uniformly categorizes them with well-defined boundaries but also effectively summarizes these hybrids, which arise from MRRs based on different criteria.
For example, historical hybrid MRRs that combine lattice-type and chain-type \cite{murata2002m,jorgensen2004modular,zykov2005self,salemi2006superbot,yim2007towards}, as per our tripartite framework, feature multiple monogamous connectors, and the modules exhibit joint DoF.
As another example, a hybrid-type MRR that combines mobile-type and chain-type includes both spatial DoF and joint DoF\cite{brown2002millibot,mondada2004swarm,guanghua2006realization,granosik2005omnitread}.
In yet another example, a hybrid-type MRR combining lattice-type and mobile-type MRRs features multiple monogamous connectors and simultaneously exhibits spatial DoF\cite{romanishin2013m,paulos2015automated,saldana2018modquad,wang2019roboat}.
In a more complex scenario, a hybrid-type MRR incorporating lattice-type, chain-type and mobile-type, under our tripartite framework, possesses multiple monogamous connectors and exhibits both spatial DoF and joint DoF\cite{wei2010sambot,kutzer2010design,davey2012emulating,liedke2013collective,belke2017mori}.
As mentioned earlier in Section \ref{sec:intro}, historical classifications of MRRs relied on various criteria related to morphology and capability, often leading to certain MRRs being classified into multiple types simultaneously. The ambiguity stemming from this situation has necessitated the adoption of a hybrid-type classification due to the absence of a precise taxonomy, consequently complicating and confusing the classification of MRRs.
The tripartite framework proposed in this paper addresses the problem by identifying key elements and classifying them, offering an additional technical perspective that enhances our understanding of MRR and enables accurate descriptions of previously uncertain cases involving hybrid-type classifications.
We hope it serves as a reference for the establishment of a recognized MRR taxonomy.

\subsubsection{\textbf{Truss}}
Truss-type MRR system consists of interconnectable beams, nodes, or struts, forming a morphology that resembles a truss.
By strategically arranging beams, nodes, or struts, truss structures provide strong structural attributes, allowing truss-type MRRs to effectively distribute bending moments and shear forces. This makes them well-suited for robust load-bearing applications and supportive structures\cite{spinos2017variable}.
Furthermore, in contrast to conventional MRRs with chain configurations, truss-type MRRs uniquely create a parallel truss mechanism with multiple DoF, capable of generating parallel directional forces akin to parallel robots\cite{tu2022freesn}.
Notably, in series MRR configurations, system performance depends on the function of the miniature modules, whereas the parallel nature of truss-type MRRs enhances the superposition effect of module capabilities, providing a novel solution that partially mitigates the limitations of series configurations.
Tetrobot\cite{hamlin1996tetrobot} was the earliest example of the truss-type MRR, and there has been a recent diversification in the design of truss-type MRRs\cite{yu2008morpho,lyder2008mechanical,spinos2017variable,tu2022freesn}. All these truss-type MRR systems can be classified as having two heterogeneous modules: one serving as truss nodes and the other as truss links.
Additionally, these systems feature joint DoF, enabling truss reconfiguration and locomotion. Some examples include spherical joint DoF between nodes and links\cite{hamlin1996tetrobot,tu2022freesn,lyder2008mechanical}, while others exhibit prismatic joint DoF  between nodes\cite{yu2008morpho,spinos2017variable}.
In general, within our proposed tripartite framework, these examples of truss-type MRRs are classified as MRR systems with heterogeneity and joint DoF.

\subsubsection{\textbf{Freeform}}
Freeform-type MRR modules emphasize their connecting capabilities, enabling connections between modules to be established from all positions and directions. Circular and spherical mechanisms are often employed to achieve this unlimited range of connection characteristics.
Freeform-type MRR enables alignment-free connections among modules, significantly enhancing fault tolerance and adaptability.
Furthermore, the flexibility of freeform connections enables diverse configurations, effectively addressing various tasks and challenges, demonstrating exceptional versatility, functionality, and scalability.
The alignment-free and flexible connection properties of freeform-type MRR lead to intriguing control strategies.
As an illustration, FireAnt's robust alignment-free connectors underpin an algorithm\cite{swissler2022reactivebuild} that leverages these connections through climbing companions, allowing robot swarms to construct a variety of adaptive structures.
For another example, FreeBOT incorporates arbitrary connection points, introducing a novel dexterous joint\cite{zong2022kinematics} class that supports highly flexible configurations and behaviors.
Within our proposed tripartite framework, freeform-type MRRs are classified as having polygamous connectors. 
Although these freeform MRRs initially incorporated circular or spherical connectors primarily to enhance connection performance, including fault tolerance and diversity, these freeform connector designs inadvertently resulted in one-to-many connection characteristics, aligning with our definition of polygamous connectors.
Slimebot\cite{tokashiki2003development} is the earliest freeform-type MRR, employing Velcro connectors arranged in a circular layout within a 2D workspace, enabling connections between modules from any position and direction on the plane.
Subsequent advancements in freeform-type MRR have delved into the application of various technologies and improved connector performance\cite{swissler2018fireant,saintyves2023self,kirby2007modular}. Recently, freeform-type MRR has undergone rapid development, extending into 3D space and giving rise to a series of 3D freeform-type MRR systems\cite{swissler2020fireant3d,liang2020freebot,tu2022freesn,zhao2022snailbot,swissler2023fireantv3}.
Through the tripartite framework proposed in this article, we strive to decode these free-form-related connector characteristics, revealing a convergence between traditional freeform classification and the concept of polygamous connectors we have introduced.

\begin{table*}[p]
  \centering
  \caption{The progression of MRR, encompassing classification, elements, technology, capability, and parameter.}
  \includegraphics[width=\linewidth]{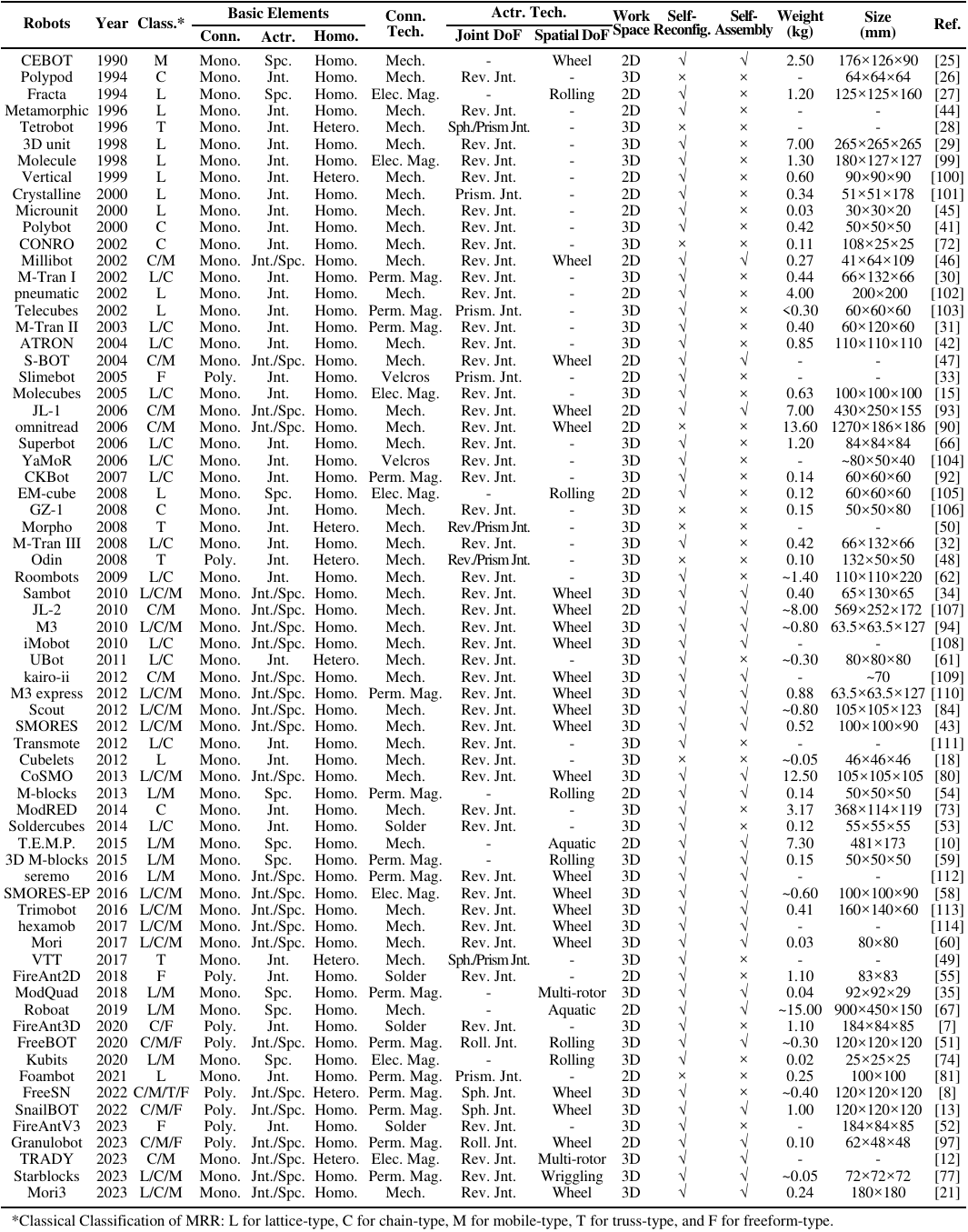}
  \label{longtable}
  
  \vspace{50pt}
\cite{fukuda1990cellular}
\cite{yim1994new}
\cite{murata1994self}
\cite{pamecha1996design}
\cite{hamlin1996tetrobot}
\cite{murata19983}
\cite{kotay1998self}
\cite{hosokawa1999self}
\cite{rus2000physical}
\cite{yoshida2000micro}
\cite{yim2000polybot}
\cite{castano2002conro}
\cite{brown2002millibot}
\cite{murata2002m}
 \cite{inou2002development}   
\cite{suh2002telecubes}
\cite{kurokawa2003m}
\cite{jorgensen2004modular}
\cite{mondada2004swarm}  
\cite{shimizu2005slimebot}
\cite{zykov2005self}
\cite{guanghua2006realization}
\cite{granosik2005omnitread}   
\cite{salemi2006superbot}
\cite{moeckel2006exploring}    
\cite{yim2007towards}
\cite{an2008cube}
\cite{zhang2008development}
\cite{yu2008morpho}
\cite{kurokawa2008distributed}
\cite{lyder2008mechanical}
\cite{sproewitz2009roombots}
\cite{wei2010sambot}
\cite{wang2010jl}
\cite{kutzer2010design}
\cite{ryland2010design}
\cite{zhao2011new}
\cite{pfotzer2012biologically} 
\cite{wolfe2012m}
\cite{russo2012design}
\cite{davey2012emulating}
\cite{qiao2012design}
\cite{cubelets}
\cite{liedke2013collective} 
\cite{romanishin2013m}
\cite{baca2014modred}
\cite{neubert2014self}
\cite{paulos2015automated} 
\cite{romanishin20153d}
\cite{bie2016systems} 
\cite{tosun2016design} 
\cite{zhang2016modular}
\cite{ch2017hexamob} 
\cite{belke2017mori} 
\cite{spinos2017variable}
\cite{swissler2018fireant}
\cite{saldana2018modquad}
\cite{wang2019roboat}
\cite{swissler2020fireant3d}
\cite{liang2020freebot}
\cite{hauser2020kubits}
\cite{ceron2021soft}
\cite{tu2022freesn}
\cite{zhao2022snailbot} 
\cite{swissler2023fireantv3}
\cite{saintyves2023self}
\cite{sugihara2023design}
\cite{zhao2023starblocks}
\cite{belke2023morphological}

\end{table*}

\section{From Elements Technologies to MRR Capabilities} \label{sec:tech2cap}
In previous sections, we mentioned that the MRR module is comprised of two inherent elements: connectors and actuators. These foundational elements shed light on essential MRR attributes, particularly with regard to topology.
When designing an MRR, the selection of core elements can significantly influence the overall performance and functionality of the system. Diverse technologies contribute to these fundamental elements, and the elemental properties evolve as technology advances.
Hence, it's essential to thoroughly assess element functionality and performance.
In this section, we track the historical evolution of MRR using the proposed tripartite framework, with a focus on the technological and performance advancements of connectors and actuators, as illustrated in Table \ref{longtable}.
The table chronicles the evolution of MRR elements over years, noting changes in attributes, capabilities, and parameters.
We also investigate the functional and performance aspects of MRR at the table, delving into technology choices for element implementation, scrutinizing its historical evolution, and discussing future designs while considering essential selection factors.
Through these analyses and summaries, our goal is to elucidate the evolution, motivations, and technologies of MRR, offering valuable references for future development.

\subsection{Connector Technology}
\begin{figure}
  \centering
  \includegraphics[width=0.8\linewidth]{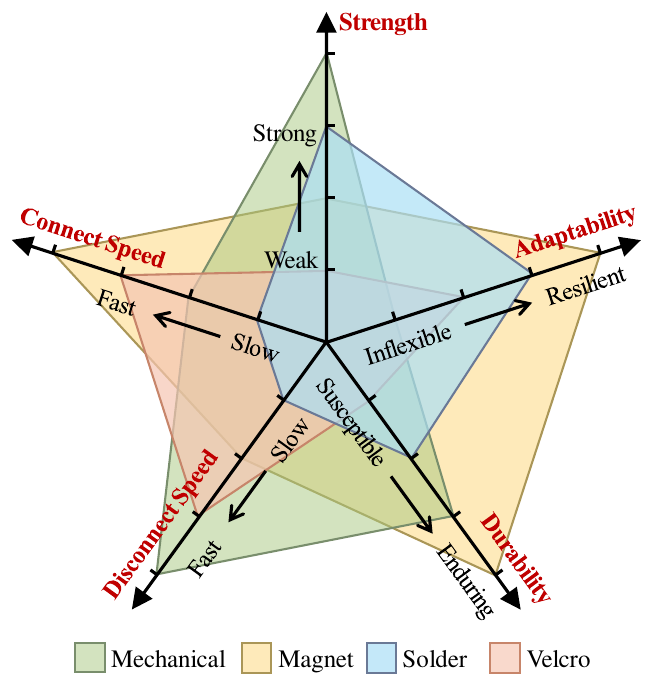}
  \caption{Analyzing various connector technologies, we can depict their characteristics on a radial scale emanating from the center of the diagram. These attributes include connection strength (weak to strong), connection speed (slow to fast), disconnection speed (slow to fast), connector durability (vulnerable to enduring), and connection adaptability (inflexible to resilient).}
  \label{fig:radar}
\end{figure}

Connectors play a crucial role as intermediaries between modules, enabling the establishment of connections within the MRR system.
When selecting a connector technology, designers must strike a balance between factors such as connection strength, adaptability, durability, connection/disconnection speed, and more  \cite{brunete2017current}, in order to attain robustness, ease of assembly, reconfiguration precision, and mechanical/electrical compatibility. Various connection technologies for MRR, including mechanical, magnetic, soldering, and Velcro connectors, each offer distinct characteristics and advantages, thereby imparting specific attributes to MRR systems. We have analyzed the performance of these connector technologies with regard to these factors, and a comparative visualization is presented in Fig. \ref{fig:radar}.
The choice of suitable connector technology ensures effective and reliable connections within the MRR system, thereby enhancing the overall performance and functionality of the robot.

\subsubsection{\textbf{Mechanical Connector}}
Mechanical connectors initiate an attractive force between modules by harnessing mechanical force or torque, representing a widely employed technology\cite{saab2019review}.
These connectors can be classified into several subtypes, including pin-and-hole\cite{castano2002conro,yim2000polybot,brown2002millibot}, hooks\cite{jorgensen2004modular,sproewitz2009roombots,wei2010sambot}
, as well as lock-and-key 
\cite{rus2000physical,qiao2012design}, each characterized by its distinct design iteration.
Regardless of design specifics, the core principle remains the same: driving the mechanical structure, pairing modules, and creating a vital attraction.
The durability, stability, and reliability of mechanical connectors primarily rely on the strength of their materials, making them generally possess robust connection strength.
Nevertheless, mechanical connectors are not well-suited for fault-tolerant docking\cite{eckenstein2017modular}, and MRRs that employ mechanical connectors often necessitate extra planning and adjustments during the connection process\cite{liu2023smores}. This can lead to elevated docking costs, including extended connection durations and increased complexity, which may pose challenges in terms of ease of use.

\subsubsection{\textbf{Magnetic Connector}}
Magnetic connectors are another commonly adopted technology in MRR. They utilize magnets to create an attractive force between modules, forming magnetic connections between permanent magnets\cite{saldana2018modquad,romanishin20153d,murata2002m}, electromagnets\cite{sugihara2023design,hauser2020kubits,tosun2016design,zykov2005self}, or magnets and ferromagnetic materials\cite{liang2020freebot,tu2022freesn,zhao2022snailbot}.
While specific implementation, design, and mechanism details may vary, the fundamental principle remains consistent: these connectors utilize magnetic fields to attract other modules, facilitating pairing and establishing connections through magnetic attraction.
Magnetic connectors are widely embraced in the MRR field for their inherent advantages in enabling rapid, fault-tolerant docking between modules through magnetic automatic mating\cite{dokuyucu2023achievements,saab2019review}. This connector type is regarded as having the highest level of fault tolerance and adaptability.
However, it's also important to consider the drawbacks of magnetic connectors. Firstly, the magnetic connection force is relatively weak, which can limit the overall robustness and stability of the robotic system. Secondly, magnetic fields in proximity to connectors may interfere with other electronic components in the MRR, necessitating careful attention to system electromagnetic compatibility.

\subsubsection{\textbf{Solder Connector}}
Solder connectors establish a connection between components by applying heat to melt solder material, utilizing a heat source to raise the solder joint temperature, which, upon cooling and solidification, forms a mechanically stable connection.
Typical materials used in MRR solder connectors include carbon-infused conductive plastic (PLA)\cite{swissler2018fireant,swissler2020fireant3d} and tin-lead mixtures\cite{neubert2014self}.
Solder connectors provide substantial connection strength while retaining flexibility, demanding less precise alignment compared to traditional mechanical connectors. Furthermore, the conductive properties of solder material allow for the integration of power and communication functions within the MRR.
Nevertheless, certain drawbacks warrant consideration. Firstly, this connection technique requires an extended heating and cooling duration, resulting in prolonged connection times. Secondly, solder material might transfer between modules during connect-disconnect cycles, potentially affecting long-term durability. Lastly, solder connectors rely on a consistent power supply to maintain the connection, posing challenges related to power management.
However, it's crucial to emphasize that the aforementioned comparisons are solely made at the technology level. When we delve into the specifics, certain smart connector designs employing solder technologies exhibit good performance, such as fault tolerance\cite{swissler2020fireant3d,swissler2023fireantv3}. Although solder connectors have seen limited implementation in MRR designs, they have displayed remarkable capabilities in multiple aspects. These works effectively illustrate the potential of solder technology for MRR applications.

\subsubsection{\textbf{Velcro Connector}}
Velcro connectors use Velcro material to establish a temporary bond through a mechanical interlock when surfaces come into contact, enabling connections between MRR modules.
Velcro connectors excel in facilitating quick and easy connections, allowing for instantaneous contact-based module connections without the requirement for additional actuator reliance\cite{shimizu2006development,moeckel2006exploring,shimizu2005slimebot}.
Furthermore, Velcro connectors provide a cost-effective connection, contributing to the affordability of the MRR.
Nevertheless, it is imperative to recognize certain constraints related to Velcro connectors. The connection strength depends on Velcro material characteristics, which could limit their suitability for high-load scenarios that demand substantial weight-bearing capacities. Furthermore, concerns about longevity arise due to the susceptibility of Velcro straps to wear, necessitating regular maintenance to ensure the ongoing operational integrity of the connection.
Despite these limitations, Velcro connectors offer a practical and cost-effective alternative for specific MRR applications, being the preferred choice in certain cases.

\begin{figure}
  \centering
  \includegraphics[width=.95\linewidth]{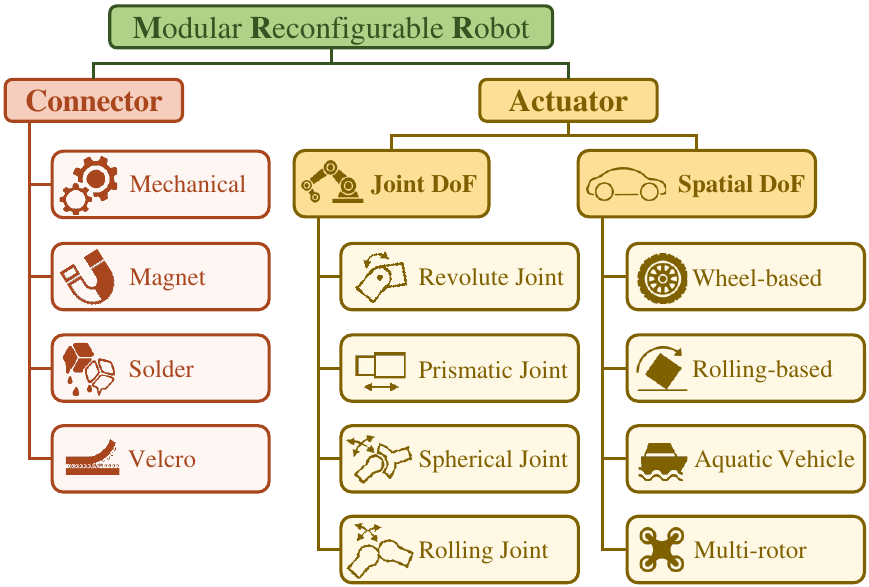}
  \caption{Various techniques implement MRR's elements. Connector methods encompass mechanical, magnetic, solder, and Velcro. Actuator technology classifies into joint DoF (e.g., revolute, prismatic, spherical, rolling) and spatial DoF (e.g., wheeled, rolling, aquatic, aerial) actuators.}
\end{figure}

\subsection{Actuator Technology} \label{sec:act tech}
The actuator serves as the pivotal element providing DoF within MRR. As mentioned earlier within the proposed tripartite framework, this DoF categorization encompasses both joint DoF and spatial DoF.
Consequently, in order to effectively attain the envisioned DoF inherent to these modules, a variety of actuator technologies for MRR have undergone extensive exploration and investigation.
These actuators that provide DoF can be categorized as joint actuators and spatial actuators. They offer joint DoF and spatial DoF, respectively, with each exhibiting a unique technological subtype and demonstrating its own significant properties; or even a combination of them, providing both types of freedom within one module by equipping it with either multiple or a single type of actuator.

\subsubsection{\textbf{Joint Actuator}}
The joint actuators within the module enable the MRR to attain joint-like rotational or angular motion by providing joint DoF.
Various joint actuator implementations arise from specific use cases. While several methods for joint actuation, such as electric motors, hydraulics, and pneumatics, have been proposed\cite{yang2018grand}, to the best of our knowledge, electric motors are currently the only employed joint actuators in MRR systems.
Electric motors are favored for their efficiency, controllability, and easy integration with modern electronics. They offer precise speed and position control, along with sufficient power output to meet the requirements of MRRs.
While electric motors are universally chosen for joint actuation within MRR systems, there is considerable diversity in the types of joints used, including revolute\cite{yim1994new,belke2023morphological,davey2012emulating,sproewitz2009roombots}, spherical\cite{spinos2017variable,tu2022freesn,zhao2022snailbot}, rolling\cite{liang2020freebot,saintyves2023self,zong2022kinematics}, and prismatic\cite{ceron2021soft,lyder2008mechanical,suh2002telecubes,yu2008morpho} joints.
The selection and design of the joint have an impact on system functionality. Depending on the type of joint, well-informed decisions take into account factors such as motion requirements, load capacity, range of motion, environmental conditions, as well as cost and manufacturing limitations. For example, revolute joints, which have been widely developed, offer better overall performance\cite{belke2023morphological,daudelin2018integrated}, while prismatic joints are better suited for specific scalable deformation tasks\cite{spinos2017variable}. Spherical joints provide a greater range of motion\cite{tu2022freesn}, and rolling joints enhance dexterity\cite{zong2022kinematics}, among other considerations.
Joint actuators enhance the adaptability and task versatility of MRRs, making them suitable for a broad spectrum of applications, including leg-like movements, manipulation tasks, and mobility scenarios in diverse environments.
Furthermore, integrated joint encoders enable precise state estimation between connected modules, enabling controlled complex motions and accurate inter-module coordination in MRRs.
Nevertheless, it's crucial to acknowledge certain limitations. Joint actuators primarily facilitate joint-like motion and lack independent movement capabilities, which can weaken multi-robot features and limit the potential for self-reconfiguration or self-assembly within MRRs.
In conclusion, the array of available joint actuator technologies presents exciting prospects for the advancement of MRR. We should actively explore the intersection of existing joint technologies and modular robotics, fostering innovative solutions and the progress of MRR actuator systems.

\subsubsection{\textbf{Spatial Actuator}} 
The spatial actuator encompasses a spectrum of propulsion mechanisms that endow modules with spatial DoF, enabling them to move individually within expansive spatial domains.
The specific implementations of spatial actuators vary according to different workspaces and operational contexts.
A significant majority of MRRs equipped with spatial actuators have spatial DoF for motion within a two-dimensional plane. These modalities further diverge depending on the specific operational environment. Land-based MRRs are predominantly wheeled vehicles\cite{fukuda1990cellular,wei2010sambot,davey2012emulating,belke2017mori} , with a smaller subset being rolling-based method\cite{romanishin2013m,liang2020freebot,hauser2020kubits}. In contrast, waterborne MRRs are primarily aquatic vehicles\cite{paulos2015automated,wang2019roboat}.
Some MRRs attain spatial DoF even within three-dimensional workspaces, enabled by multi-rotor\cite{saldana2018modquad,sugihara2023design}.
With the capability to move independently, MRR modules can alter their spatial distribution\cite{liu2023smores}, effectively reshaping the connections between modules to accommodate dynamic environmental conditions and mission requirements. This mobility capability provides MRR modules with adaptability in terms of configuration and personalized maneuverability, thereby facilitating navigation and exploration in expansive environments.
Notably, the spatial DoF of each module are determined by its specific propulsion mechanism, signifying that the relative motion between modules also depends on the spatial actuator of each module.
Consequently, MRRs equipped only with such actuators lack the necessary force output for effective interaction between modules. This limitation restricts their capabilities for locomotion and manipulation.
In summary, these remarkable technologies furnish spatial actuators for MRRs. Concurrently, the study of these mobile technologies provides a reference for us: actively exploring existing robotic mobility technologies, along with considering their intersection with modular robotic technologies, holds the potential for innovative solutions to enable individual mobility in MRRs.

\subsubsection{\textbf{Integrated Actuator}}
In the previous discussion, the two actuators offer distinct DoF: spatial actuators provide DoF for self-movement, while joint actuators provide DoF for joint motion.
These two DoF are non-conflicting and can be synergistically combined to enhance MRR capabilities.
Recognizing the inherent benefits of combining both types of actuators, certain MRR designs incorporate these two actuator types to leverage their respective strengths and maximize advantages.
For instance, certain MRR systems combine wheel drives for individual ground mobility with revolute joint actuators for joint motion among modules\cite{wei2010sambot,davey2012emulating,kutzer2010design,russo2012design,liedke2013collective}.
For another instance, the recent TRADY\cite{sugihara2023design} enables modules to move independently through the air using multi-rotors and also incorporates joint actuators to facilitate chain motion after module connection.
These MRRs have two distinct types of actuators: joint actuators and spatial actuators. The fusion of these two actuators provides both DoF, enhancing MRR performance.
Furthermore, recent advancements are focused on the integrated actuators, where a single actuator assumes the dual function of providing both spatial DoF and joint DoF.
Mori's\cite{belke2017mori,belke2023morphological} actuator is positioned at the edge and is equipped with friction wheels, allowing independent movement through coordinated edge rotation. When modules are linked at the edge, this rotation triggers a chain-like motion.
The recent FreeBOT\cite{liang2020freebot} and SnailBOT\cite{zhao2022snailbot} are capable of independent land locomotion, as well as surface traversal on peer modules, resulting in joint-like motion.
The integrated actuator combines two types of actuators, providing spatial DoF and joint DoF, which enhances the MRR function. However, due to its dual role, it inevitably suffers from performance and professionalism limitations. Current integrated actuators have constraints in joint strength, output force, and stability, necessitating ongoing research to improve their future performance.

\subsection{MRR Capabilities Evolution}
\begin{figure*}
  \centering
  \includegraphics[width=.86\linewidth]{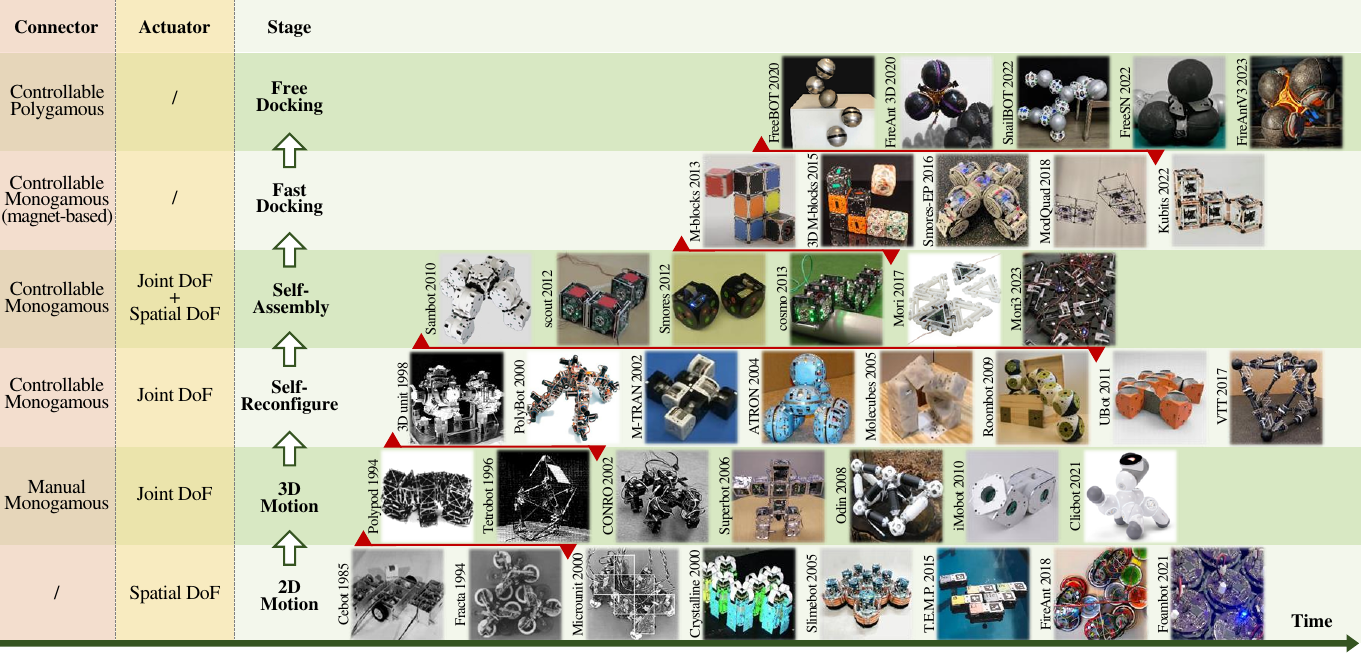}
  \caption{The correlation between element functions and MRR capabilities. It outlines the sequential advancement of MRR functionalities across six key stages: 2D motion, 3D motion, self-reconfiguration, self-assembly, fast docking, and free docking.}
  \label{fig:stage}
\end{figure*}

The MRR system has consistently demonstrated significant advancements over time, as depicted in Table \ref{longtable}. Starting in a 2D environment, technological progress has enabled MRR to extend its capabilities into a 3D environment, encompassing advanced functions such as self-reconfiguration, self-assembly, and enhanced connectivity. This section delves into the developmental milestones of MRR, with a focus on the six key stages illustrated in Fig. \ref{fig:stage}. Each stage symbolizes the technological and design innovations of its respective era, underscored by MRR's ability to adapt to growing mission requirements.

\subsubsection{\textbf{2D Motion}}
During the early stages of MRR, the primary focus was on 2D motion.
CEBOT\cite{fukuda1990cellular} pioneered the concept of combining individual robots to form a planar mobile multi-robot system with physical connectors, laying the foundation for MRR exploration.
Within the proposed tripartite framework's scope, the MRRs in this stage are equipped with actuators that provide spatial DoF, enabling them to self-move within the 2D spatial domain.
Furthermore, during this phase, the connectors were predominantly monogamous and manually connected\cite{murata1994self,yoshida2000micro,rus2000physical,shimizu2005slimebot,paulos2015automated}.
Nevertheless, limitations on 2D locomotion restrict the wide-ranging applicability of these robots. At this stage, challenges associated with limited mobility and spatial awareness hinder their potential application in various domains.

\subsubsection{\textbf{3D Motion}}
Building upon the concepts established in the initial stage, the researcher initiated the development of an MRR capable of operating within a 3D environment.
The introduction of a MRR engineered for 3D spatial operations represents a significant milestone, expanding the range of potential uses and enhancing the prospective applications of MRR.
The Polypod\cite{yim1994new} is the pioneering MRR capable of 3D movements, featuring modules that can be arranged into various chain or branch-like morphologies, enabling a wider range of task-related functionalities.
Following that, certain MRRs\cite{hamlin1996tetrobot,castano2002conro,salemi2006superbot,lyder2008mechanical,ryland2010design} pushed the field forward by presenting innovative design paradigms, unique movements, and multifaceted functionalities.
During this stage, the MRR evolved alongside the chain-type MRR, featuring actuators that provided joint DoF, while manual monogamy connectors continued to dominate.
Once these chain-type modules are interconnected with connectors, the joint actuators facilitate movement in 3D space.
Nonetheless, manual assembly and limited reconfiguration capabilities remain significant obstacles that MRR relies on, hindering their flexibility and restricting their potential applications.

\subsubsection{\textbf{Self-Reconfiguration}}
The development of self-reconfigurable MRRs, which can self-alter their connectivity to assume various configurations, represents a significant milestone in the field.
The 3D unit\cite{murata19983} represents the inaugural self-reconfigurable MRR system, characterized by its 3D matrix structure, wherein individual units possess the ability to autonomously rearrange themselves to achieve diverse spatial configurations.
The capacity for self-reconfiguration has captivated the attention of the MRR community for a considerable duration\cite{yim2000polybot,murata2002m,jorgensen2004modular,zykov2005self,spinos2017variable}. 
The emergence of this phase of MRR coincided with the introduction of hybrid-type MRR systems, where specific joint actuators coexist with monogamous connectors.
In contrast to the previous stage, the emphasis of this stage is on establishing repeatable and controllable connections using connectors that facilitate the connection process, while also enabling the joint actuators to reposition them for reconfiguration.
This stage greatly enhances the versatility and adaptability of MRRs for various tasks and environments. Self-reconfiguring robots exemplify this enhanced adaptability, opening the door to new dynamic applications and more functional robotic systems.
Nevertheless, existing limitations persist; the primary challenge in this stage concerns the restricted scalability of MRR, as the robot's size is pre-determined by the initial module count and cannot be increased by adding new modules.

\subsubsection{\textbf{Self-Assembly}}
Following the advancement in self-reconfiguration, MRR systems were subsequently engineered to incorporate self-assembly functionalities. Self-assembly means that modules connect to each other from a previously unconnected state through their own mobility.
The Sambot\cite{wei2010sambot}, being the first MRR with self-assembly capabilities, is equipped with wheel drive for independent movement within a 2D plane, and its operational domain extends to three dimensions after assembly.
Following its introduction, the capacity for self-assembly in MRR has attracted sustained academic interest, with these MRR exhibiting increased complexity and multifunctional movement after assembly\cite{russo2012design,davey2012emulating,liedke2013collective,belke2023morphological}.
At this stage, the MRR landscape is still primarily characterized by hybrid MRRs that incorporate controllable monogamous connectors based on mechanical/magnetic technology. Nevertheless, there is also a renewed focus on individual mobility, achieved by outfitting each module with joint actuators and spatial actuators, or even integrated actuators. Each module possesses both spatial DoF and joint DoF.
In contrast to the previous stage, the incorporation of spatial DoF enables the MRR to move independently within a vast workspace. This capability allows the MRR to modify its configuration by repositioning its modules at different locations, and the system can be expanded through the addition of new modules.
The self-assembly process is frequently slow and inefficient, which imposes fresh demands on module docking and introduces novel challenges to the field as it merges discrete modules into a entity.

\subsubsection{\textbf{Fast Docking}}
As MRR technology advances, connectivity performance has become a central focus in the field, motivating our efforts to design MRR systems capable of rapid interconnection, thus improving the efficiency of self-assembly.
SMORES-EP\cite{tosun2016design} builds upon its predecessor\cite{davey2012emulating} by introducing an enhanced EP face connector, a controllable connector that utilizes electromagnetic coil coupling to ensure fast and fault-tolerant connections between modules.
A series of MRRs demonstrated rapid self-assembly capabilities through innovative connector designs, signifying a substantial improvement in self-assembly efficiency\cite{romanishin2013m,romanishin20153d,saldana2018modquad,hauser2020kubits}.
At this stage, enhancements primarily focus on performance. While MRRs remain primarily hybrid-type and continue to employ monogamy connectors, the speed of these connectors has gained prominence because of their direct impact on reconfiguration and assembly performance. By primarily leveraging magnetic technology and innovative design, the time needed to establish connections for MRRs has been significantly reduced.
In this stage, the MRRs incorporate a fast-connect mechanism, enhancing the speed and versatility of the MRR system. As a result, this stage represents a significant increase in self-assembly efficiency, which enhances the feasibility of MRRs for time-critical operations. The accelerated docking process enables quicker reconfiguration and self-assembly, expanding the range of practical real-world applications.

\subsubsection{\textbf{Free Docking}}
At this stage, although connection performance remains crucial, there has been a recent shift in focus towards simplifying the process of establishing connections and enabling the connection of MRRs in any position, a concept known as free docking.
The classification of freeform-type MRR has a long history\cite{shimizu2005slimebot}, but only recently has there been a shift in focus towards its operation in three-dimensional space.
Around the same time in 2020, both FreeBOT\cite{liang2020freebot}  and FireAnt3D\cite{swissler2020fireant3d} were introduced as the initial wave of MRRs capable of 3D freeform connections. They utilized magnetic and fusion technologies, respectively, to achieve versatile, polygamous connectors.
Following that, a series of freeform-type robots have been introduced\cite{tu2022freesn,zhao2022snailbot,swissler2023fireantv3,saintyves2023self}, each of them possessing the capability for free docking in three dimensions.
At this stage, connection fault tolerance prompts innovative connector designs for polygamous connectors in 3D space, thereby altering the foundational topology characteristics of MRRs and the rate of connection fault tolerance.
This development enhances the adaptability and versatility of MRRs, resulting in MRRs that are more flexible, offering expanded potential applications and stimulating the evolution of designs that are increasingly functional and adaptable.

\section{Summary and Outlook}\label{sec:summary}
MRRs have attracted significant attention because of their unique ability to be reconfigured into various shapes and configurations. This has inspired researchers to explore a wide range of designs, technologies, functions, and applications, thereby continuously advancing the field.
Nonetheless, the varied research efforts within different teams shape field trends and factions. The absence of standardized MRR classification and definitions leads to confusion, contributing to a somewhat chaotic state in the MRR field.
To address these misunderstandings and uncertainties, this paper presents a novel tripartite framework that decodes MRR into three essential elements: connectors, actuators, and homogeneity.
Additionally, this paper conducts a comprehensive analysis of the evolving technologies in MRR development, thereby enhancing our comprehension of its distinct evolutionary stages.
The proposed tripartite framework and its basic elements provide a complementary perspective at a technical level, enhancing the understanding of MRR hardware.
While MRR has achieved considerable advancements, there remain unresolved challenges and promising avenues for research. This underscores the need for ongoing research efforts to augment MRR capabilities and unlock its full potential across diverse applications, ultimately propelling the field of robotics forward.

\subsubsection*{\textbf{Connector}}
Over the years, a diverse range of connector designs and technologies have emerged\cite{dokuyucu2023achievements,saab2019review}, demonstrating the continuous development of the field.
The evolution of connector design began with ensuring reliable module connections\cite{yim1994new,zykov2005self,wei2010sambot}, transitioned to facilitating efficient reconfiguration\cite{tosun2016design,romanishin2013m,saldana2018modquad}, and now places a greater emphasis on adaptable connectors\cite{liang2020freebot,tu2022freesn,swissler2020fireant3d} to meet diverse application demands.
Despite the significant advancements in connector technology, several challenges and unresolved research questions still persist regarding its design and implementation.
Amidst the emergence of novel technologies, accompanied by their inherent features and capabilities, they simultaneously introduce trade-offs and restrictions; the primary challenge involves achieving a balance between the demands for strength and the requirement for fast, fault-tolerant efficiency.
At present, there is no connector design that comprehensively combines speed, strength, and fault tolerance, which has spurred ongoing research efforts to develop connectors that excel in all of these aspects.

\subsubsection*{\textbf{Actuator}}
Much like the challenges encountered in actuator technology for legged robots\cite{yang2018grand}, achieving a balance between superior performance, compactness, and weight in actuator designs for MRRs poses a similar challenge.
In the realm of MRR, this issue is further intensified, as functionality demands tight integration of all elements, including actuators, within the smallest possible module.
The challenges posed by the miniaturization of actuators place significant constraints on the performance of MRRs.
Balancing the reduction in actuator size with an increase in output force within MRR modules presents a challenge, necessitating trade-offs between size and performance. Future advancements in actuator technology depend on improvements in motor capabilities, materials science, and energy efficiency.
Furthermore, dynamics are often overlooked in MRR design, with researchers primarily focusing on kinematics, a tendency attributed to the current MRR actuators' limited actuation performance that constrains the exploration of dynamic behaviors.
Dynamics constitute a fundamental and captivating aspect of MRR, encompassing the interplay of forces, torques, and energy between diverse modules and their surroundings. The dynamics of MRRs are poised to receive increased research attention as forthcoming advancements in actuator technologies unfold.

\subsubsection*{\textbf{Homogeneity}}
Historically, the primary focus of MRR research has been the development of homogeneous modules that are identical and replicable\cite{murata2002m,sproewitz2009roombots,davey2012emulating,belke2017mori}, aiming to improve fault tolerance and simplify the economical replacement of failed modules.
However, the compact nature of these modules frequently impacts their performance due to the integration of connectors, actuators,  controllers, batteries, and sensors, leading to heightened weight and  diminished capability.
Similar to the barrel principle, when utilizing homogeneous MRRs in system applications, the reduced performance of a single module invariably affects the maximum capability of the overall system.
To address this issue, we can draw inspiration from general modular robotics, where diverse module designs enable specialization in function and performance, thereby facilitating efficient task distribution within the system.
For example, consider the modular robotic arm UR5\cite{ur5}, which consists of various modular components for its shoulder, elbow, and wrist joints. Each of these components exhibits unique performance characteristics and weight, consequently enhancing the overall system's efficiency and performance.
Exploring heterogeneous design principles in MRR holds the promise of advancing the field. By incorporating various specialized modules, it becomes possible to engineer multifunctional and efficient MRR systems boasting heightened performance.

%

\bibliographystyle{IEEEtran}
\bibliography{IEEEexample}

\end{document}